\documentclass[11pt]{article}

\usepackage[utf8]{inputenc}
\usepackage[T1]{fontenc}
\usepackage{times}
\usepackage[margin=1in]{geometry}
\usepackage{xcolor}
\usepackage{setspace}
\usepackage{graphicx}
\usepackage{booktabs}
\usepackage{array}
\usepackage{longtable}
\usepackage{multirow}
\usepackage{enumitem}
\usepackage{url}
\usepackage[
    colorlinks=true,
    linkcolor=blue,
    citecolor=blue,
    urlcolor=blue,
    filecolor=blue
]{hyperref}

\usepackage{amsmath,amssymb}
\usepackage{algorithm}
\usepackage[noend]{algorithmic}
\usepackage{color}
\usepackage{xspace}
\usepackage{natbib}
\usepackage{titling}

\title{Large Language Models for Assisting American College Applications}

\author{%
Zhengliang Liu$^{1*,\dagger}$, Weihang You$^{1*}$, Peng Shu$^{1*}$, Junhao Chen$^{1*}$, Yi Pan$^{1*}$, Hanqi Jiang$^{1*}$, \\
Yiwei Li$^{1*}$, Zhaojun Ding$^{1*}$, Chao Cao$^{2*}$, Xinliang Li$^{1*}$, \\
Yifan Zhou$^{1*}$, Ruidong Zhang$^{1*}$, Shaochen Xu$^{1*}$, Wei Ruan$^{1*}$, Huaqin Zhao$^{1*}$, \\
Dajiang Zhu$^{2}$, Tianming Liu$^{1}$ \\
[0.6em]
$^{1}$University of Georgia \\
$^{2}$University of Texas at Arlington \\
\vspace{0.3em}
{\small $^{*}$These authors contributed equally.}
}

\date{}

\begin{document}

\begin{center}
{\Large\bfseries \thetitle \par}
\vspace{0.8em}
{\normalsize \theauthor \par}
\end{center}
\vspace{1.2em}

% -------------------- Abstract --------------------
\begin{abstract}
American college applications require students to navigate fragmented admissions policies, repetitive and conditional forms, and ambiguous questions that often demand cross-referencing multiple sources. 
We present \textit{EZCollegeApp}, a large language model (LLM)--powered system that assists high-school students by structuring application forms, grounding suggested answers in authoritative admissions documents, and maintaining full human control over final responses. 
The system introduces a mapping-first paradigm that separates form understanding from answer generation, enabling consistent reasoning across heterogeneous application portals. 
EZCollegeApp integrates document ingestion from official admissions websites, retrieval-augmented question answering, and a human-in-the-loop chatbot interface that presents suggestions alongside application fields without automated submission. 
We describe the system architecture, data pipeline, internal representations, security and privacy measures, and evaluation through automated testing and human quality assessment. Our source code is released on GitHub (\href{https://github.com/ezcollegeapp-public/ezcollegeapp-public}{github.com/ezcollegeapp-public/ezcollegeapp-public}) to facilitate the broader impact of this work.
\end{abstract}

% -------------------- Corresponding Author Footnote --------------------
\footnotetext[2]{Corresponding author: \href{mailto:zl18864@uga.edu}{zl18864@uga.edu}.}

\pagebreak

\tableofcontents

\pagebreak

% ======================================================
\section{Introduction}
\label{sec:intro}
Applying to college in the United States is a complex, high-stakes process that demands significant cognitive and administrative effort from high-school students and their families~\cite{giersch2018academic}. Applicants must navigate a fragmented landscape of institution-specific admissions policies, complete repetitive and conditionally structured forms across multiple platforms (e.g., Common Application, Coalition Application, university-specific portals), and respond to questions whose intent is often ambiguous without consulting external documentation. The process is further complicated by the fact that admissions requirements, deadlines, and policies vary widely across institutions, and are frequently updated without centralized notification. Students must cross-reference information from official admissions websites, frequently asked questions (FAQs), community forums, and counselor advice to ensure that their applications are complete, accurate, and strategically aligned with institutional expectations.

The recent emergence of large language models (LLMs) has opened new possibilities for automating or assisting with complex information-processing tasks, including question answering, document summarization, and form completion (\cite{lee2025multimodality,ding2024foundation,latif2024systematic,most2024evaluating,shu2024llms,tian2024assessing}). However, naïvely applying LLMs to college applications introduces significant risks. LLM models trained on broad web corpora may generate plausible but factually incorrect or outdated information, a phenomenon commonly referred to as hallucination. In the context of college admissions, such errors can have serious consequences, including application rejection, missed deadlines, or misrepresentation of a student's qualifications. Furthermore, the opacity of end-to-end neural generation makes it difficult for users to verify the correctness of model outputs or to understand the reasoning behind suggested answers, undermining trust and accountability.

EZCollegeApp is designed to address these challenges by providing grounded, transparent, and human-supervised assistance throughout the college application process via a chatbot interface. Rather than automatically filling forms or generating answers in isolation, the system adopts a human-in-the-loop approach that preserves student agency while reducing cognitive load \cite{mosqueira2023human, Zhu2018}. The core design principle is that the system should suggest answers based on explicit, verifiable evidence from authoritative sources, and that the student should retain full control over which suggestions to accept, modify, or reject \cite{AbuRasheed2024}. All suggested content is accompanied by inline citations that link back to the original source documents, enabling users to verify accuracy and understand the provenance of each answer. Critically, the system never auto-submits forms or makes decisions on behalf of the user; it operates strictly as an assistive tool that augments, rather than replaces, human judgment \cite{Schmager2025}.

% The system architecture is built around three key technical innovations. First, EZCollegeApp employs a mapping-first paradigm that decouples the understanding of form questions from the generation of answers. Rather than directly mapping raw form text to responses, the system first resolves each question to a canonical field in an internal schema that captures the semantic intent and data type constraints of the question. This abstraction layer enables consistent reasoning across heterogeneous application portals, as semantically equivalent questions (e.g., "What is your GPA?" and "Please report your cumulative grade point average") are mapped to the same canonical field and answered using the same underlying information. Second, the system integrates a comprehensive knowledge base constructed from official admissions websites, curated FAQs, and relevant community forums, with explicit source separation and provenance tracking to distinguish authoritative information from anecdotal advice. Third, the system employs retrieval-augmented generation (RAG) to ground all suggested answers in specific document chunks retrieved from the knowledge base, ensuring that responses are factually supported and traceable to their source materials.
Second, the system constructs two complementary knowledge bases: (i) an \textit{institutional knowledge base} from official admissions websites, FAQs, and community forums with explicit source separation; and (ii) a \textit{personal knowledge base} from each student's uploaded materials, indexed through hierarchical document structure extraction that preserves section boundaries and page mappings, enabling retrieval through LLM-guided tree navigation without vector embeddings \cite{Rashkin2021}. Third, the system employs an \textit{agentic retrieval} architecture where the language model autonomously decides when to access knowledge bases, invoking retrieval as a tool only when context is required rather than automatically retrieving for every query. Fourth, all answers are generated through retrieval-augmented generation with explicit inline citations linking to specific source documents and page ranges, ensuring responses are factually grounded and verifiable \cite{Nakano2021WebGPT, Zhang2023StructGPT}. Our agent also provides long-term memory so that user preferences, stable facts, and task state remain consistent across sessions rather than being lost when the short context window expires. This capability is designed for workflows that unfold over days or weeks, where continuity materially reduces repeated clarification and prevents fragmented execution. In high-stakes or high-precision scenarios, the memory layer is paired with conservative refusal behaviors when evidence is insufficient, so that the agent avoids inventing “remembered” information and instead requests confirmation or proceeds without the missing context.

This paper presents the system design, implementation, and evaluation of EZCollegeApp. We describe the end-to-end architecture, including the conversational front-end for document collection, the back-end pipeline for document processing and semantic indexing, and the browser-based Application Copilot that delivers real-time assistance during form completion. We detail the internal canonical form representation and the three-tier field mapping algorithm that enables cross-portal consistency \cite{Rahm2001}. We explain the admissions document collection and processing pipeline, including web crawling strategies, text extraction, semantic chunking, and vector database indexing \cite{Lewis2020RAG}. We describe the question-answering system for both general admissions queries and form-specific answer generation, with a focus on grounded retrieval and citation transparency \cite{Menick2022GopherCite, Rashkin2021}. Finally, we present evaluation results from automated testing and human quality assessment, demonstrating that the system achieves high fill rates (80-85\% of questions answered), strong citation validity (92\% of citations are semantically relevant), and positive user feedback on usefulness, clarity, and trustworthiness \cite{Kiela2021}.

The contributions of this work are threefold. First, we introduce a mapping-first approach to form understanding that separates semantic intent from surface-level question wording, enabling robust cross-portal reasoning. Second, we demonstrate a practical implementation of source-separated, retrieval-augmented generation for a high-stakes domain where correctness and transparency are paramount. Third, we provide a comprehensive system description and evaluation that can serve as a reference for future work on LLM-assisted information systems in domains requiring human oversight and accountability.

% ======================================================
\section{Problem Setting and Design Principles}
\label{sec:problem}

The college application process in the United States presents a unique set of technical and user experience challenges that distinguish it from other form-filling or question-answering domains. Understanding these challenges is essential to motivating the design decisions underlying EZCollegeApp. This section characterizes the problem space and articulates the core design principles that guide the system's architecture and implementation.

\subsection{Challenges in American College Applications}

College application platforms differ widely in question wording, structure, and conditional logic, creating a fragmented and cognitively demanding experience for applicants. Identical concepts (e.g., coursework reporting, standardized testing policies, extracurricular activity descriptions) may appear under different names, formats, and organizational structures across platforms. Furthermore, answers to many questions depend on nuanced institutional rules and policies that are documented separately on admissions websites, often in dense prose or FAQ formats that require careful interpretation.

\subsubsection{Heterogeneous Form Structures}

Different application platforms employ distinct form schemas, field naming conventions, and data entry mechanisms \cite{Rahm2001}. The Common Application, for example, organizes questions into hierarchical sections (Profile, Family, Education, Testing, Activities, Writing) with standardized field labels and data types. In contrast, university-specific portals may use custom question wording, non-standard field types, and idiosyncratic organizational structures. For instance, one institution may ask "Please list your extracurricular activities in order of importance," while another may ask "Describe your involvement in school and community organizations," and a third may provide a structured table with separate fields for activity name, role, hours per week, and weeks per year \cite{Doan2012}. These variations make it difficult to reuse information across applications, forcing students to manually reformat and re-enter the same underlying data multiple times \cite{Norman2013}.

\subsubsection{Conditional Logic and Dynamic Fields}

Many application forms include conditional logic, where the visibility or required status of certain fields depends on the values entered in other fields. For example, if a student indicates that they are not a U.S. citizen, the form may hide the "Social Security Number" field and display a "Visa Type" field instead. Similarly, if a student selects "Yes" for "Have you taken any college courses while in high school?", the form may dynamically reveal a section for reporting dual enrollment coursework. These dependencies create a complex state space that must be navigated carefully to avoid incomplete or inconsistent applications \cite{Jackson2012}. Automated or semi-automated form-filling systems must respect these conditional rules, ensuring that suggestions are only provided for fields that are currently visible and applicable \cite{Amershi2019}.

\subsubsection{Ambiguous and Context-Dependent Questions}

Form questions are often ambiguous or context-dependent, requiring interpretation based on surrounding fields, section headings, or external documentation \cite{McCann2018}. For example, the question "When did you start?" could refer to the start date of an activity, the start date of high school, or the start date of employment, depending on the form context \cite{Yu2018Spider}. Similarly, questions like "Please describe your most meaningful experience" or "Why are you interested in this major?" are open-ended and require synthesizing information from multiple sources (e.g., personal essays, activity descriptions, academic interests) to generate a coherent and contextually appropriate response \cite{Yang2018HotpotQA}. Resolving these ambiguities requires not only natural language understanding but also reasoning about the form's structure and the user's profile.

\subsubsection{Policy and Requirement Complexity}

Admissions policies vary significantly across institutions and are subject to frequent updates \cite{HoxbyTurner2015}. For example, some institutions require SAT or ACT scores, while others have adopted test-optional or test-blind policies \cite{Belasco2015}. Some institutions accept self-reported test scores, while others require official score reports. Some institutions have specific requirements for course prerequisites (e.g., four years of math, including calculus), while others evaluate applications holistically without rigid prerequisites. These policies are typically documented on admissions websites, but the information is often scattered across multiple pages, embedded in lengthy FAQ documents, or expressed in ambiguous language that requires interpretation \cite{Russell1993}. Students must cross-reference these sources to ensure that their applications meet institutional requirements, a process that is time-consuming and error-prone \cite{Norman2013}.

\subsubsection{High Stakes and Low Tolerance for Error}

Unlike many other form-filling tasks, college applications are high-stakes and have low tolerance for error \cite{Amodei2016}. Incorrect or incomplete information can result in application rejection, missed deadlines, or misrepresentation of a student's qualifications. Furthermore, once an application is submitted, it is typically not possible to make corrections or updates, meaning that errors are permanent \cite{Holstein2019}. This high-stakes nature demands that any automated or semi-automated assistance system prioritize correctness and transparency over convenience or speed. Users must be able to verify the accuracy of suggested answers and retain full control over what information is submitted \cite{Amershi2019}.

\subsection{Design Principles}

EZCollegeApp is guided by five core design principles that address the challenges outlined above while ensuring that the system remains trustworthy, transparent, and aligned with user needs. These principles inform every aspect of the system's architecture, from data collection and processing to user interface design and deployment.

\subsubsection{Grounded Assistance}

All suggested answers generated by EZCollegeApp are derived from explicit, verifiable sources. The system does not rely on the language model's parametric knowledge or generate answers based on unsupported inference \cite{Lewis2020RAG}. Instead, every answer is grounded in specific document chunks retrieved from the user's uploaded materials (e.g., transcripts, resumes, essays) or from the system's knowledge base of admissions documentation (e.g., official websites, FAQs, community forums) \cite{Karpukhin2020}. Each suggested answer is accompanied by inline citations that link back to the source documents, enabling users to verify the factual basis of the suggestion and to assess its relevance and accuracy \cite{Menick2022GopherCite}. This grounding mechanism serves two purposes: it reduces the risk of hallucination or fabrication, and it builds user trust by making the reasoning process transparent and auditable \cite{Ji2023}.

\subsubsection{Human-in-the-Loop}

EZCollegeApp adopts a strict human-in-the-loop design, meaning that the system never submits or auto-fills responses without explicit user action \cite{mosqueira2023human}. The Application Copilot presents suggested answers as overlays or tooltips adjacent to form fields, but it does not automatically populate the fields. Instead, users must manually copy suggested content to the clipboard and paste it into the form, or type their own answers after reviewing the suggestions \cite{Amershi2019}. This deliberate friction ensures that users remain actively engaged in the application process and retain full control over what information is submitted. It also provides an opportunity for users to review, edit, or reject suggestions before they become part of the official application \cite{Shneiderman2020}. This design choice reflects the recognition that college applications are high-stakes documents that require human judgment and accountability, and that automation should augment, rather than replace, human decision-making \cite{Amodei2016}.

\subsubsection{Structured Reasoning}

EZCollegeApp employs internal canonical representations that decouple the understanding of form questions from the generation of answers \cite{Manning2008}. Rather than treating each form question as a unique, isolated input to the language model, the system first maps the question to a canonical field in an internal schema that captures the semantic intent, data type, and formatting constraints of the question \cite{Rahm2001}. This abstraction layer enables consistent reasoning across heterogeneous application portals, as semantically equivalent questions are mapped to the same canonical field and answered using the same underlying information. The canonical schema also encodes conditional logic and dependency relationships between fields, allowing the system to respect form constraints and avoid generating suggestions for fields that are not currently visible or applicable. This structured reasoning approach improves both the correctness and consistency of generated answers, and it simplifies the task of maintaining and updating the system as new application portals are added or existing portals change their form structures.

\subsubsection{Practical Deployment}

EZCollegeApp prioritizes robustness, usability, and real-world deployability over benchmark optimization or academic novelty \cite{Lipton2019}. The system is designed to operate reliably in production environments, handling diverse document formats, noisy or incomplete user inputs, and varying network conditions. The architecture emphasizes modularity and fault tolerance, with clear separation of concerns between document processing, reasoning, and user interaction \cite{Saltzer1984}. The user interface is designed to be lightweight and non-intrusive, minimizing cognitive load and avoiding disruption to the user's workflow \cite{Sweller1988}. The system also includes comprehensive error handling, logging, and monitoring to facilitate debugging and continuous improvement. Rather than aiming for perfect accuracy on a narrow benchmark task, EZCollegeApp seeks to provide consistent, useful assistance across a wide range of real-world application scenarios, with graceful degradation when the system encounters questions or documents that it cannot handle confidently \cite{Sculley2015}.

\subsubsection{Extraction Over Generation}

A fundamental design principle of EZCollegeApp is that the system exclusively extracts and reorganizes information from user-provided documents \cite{Shneiderman2020}. It does not generate new content or assist with creative writing tasks. This principle distinguishes EZCollegeApp from general-purpose AI writing assistants and reflects a deliberate philosophical stance about the role of AI in the college application process.

The system operates under a strict constraint: every piece of information presented to the user must be directly traceable to an existing source document (such as transcripts, resumes, recommendation letters, or previously written essays) or to factual admissions documentation (such as official university websites and application guidelines) \cite{Rashkin2021}. When a form question requires a factual answer, such as listing extracurricular activities, reporting academic honors, or providing contact information, the system identifies the relevant information in the user's uploaded materials and formats it appropriately for the target field \cite{Eaton2023}. When a form question asks for a written response, such as a personal statement, supplemental essay, or short-answer question, the system does not generate or draft the response. Instead, it may identify relevant excerpts from previously written essays that could be adapted by the user, or it may leave the field unmarked to signal that original writing is required.

This design choice is motivated by several considerations. First, college application essays serve as an authentic representation of the applicant's voice, perspective, and writing ability. Admissions officers use these essays to assess not only what the applicant has to say, but how they say it: their command of language, their ability to reflect critically, and their capacity for self-expression \cite{Park2014}. An essay generated by an AI system, even if based on facts from the applicant's life, would fundamentally misrepresent the applicant's own capabilities and undermine the integrity of the admissions process. EZCollegeApp is designed to support students in completing their applications efficiently and accurately, not to enable submission of work that is not genuinely their own.

Second, the skill of articulating one's experiences, goals, and values in writing is itself a learning objective of the college application process \cite{Biggs2011}. Students who engage deeply with essay prompts often gain self-awareness and clarity about their own motivations and aspirations. By refusing to generate essay content, EZCollegeApp preserves this valuable aspect of the application experience and encourages students to develop their own voice rather than relying on algorithmic substitutes.

Third, this principle aligns with evolving academic integrity standards and institutional policies regarding the use of AI tools in educational contexts. Many universities have established guidelines that distinguish between acceptable and unacceptable uses of AI assistance in application materials. While using AI to organize factual information (for example, formatting a list of activities) is generally considered appropriate, using AI to draft or substantially revise personal essays is widely regarded as a violation of academic honesty. By limiting its functionality to information extraction and reorganization, EZCollegeApp ensures that all use cases fall clearly within the bounds of ethical and permissible assistance.

Finally, this design principle provides a natural safeguard against the misuse of the system. Because EZCollegeApp cannot generate novel content, it cannot be used to fabricate experiences, exaggerate achievements, or create misleading narratives. The system can only present information that already exists in the user's documents, and any discrepancies between the suggested answers and the user's actual record will be immediately apparent to the user during the review process. This constraint reinforces the grounded assistance principle and further enhances the trustworthiness of the system.

In practice, this means that EZCollegeApp is designed to handle structured, factual form fields with high accuracy and efficiency, while gracefully declining to assist with unstructured, creative writing tasks. This division of labor allows students to benefit from automation where it is most valuable: reducing repetitive data entry and ensuring consistency across multiple applications, while preserving the authenticity and educational value of the writing process.

These five principles—grounded assistance, human-in-the-loop design, structured reasoning, practical deployment, and information extraction rather than content generation—collectively define the philosophy and approach of EZCollegeApp. They reflect a commitment to building AI systems that are trustworthy, transparent, and aligned with the needs and values of their users, particularly in high-stakes domains where errors can have significant consequences and where the authenticity of user contributions must be preserved.

\subsubsection{Security and Privacy Measures}

\begin{figure}[htbp]
    \centering
    \includegraphics[width=\textwidth]{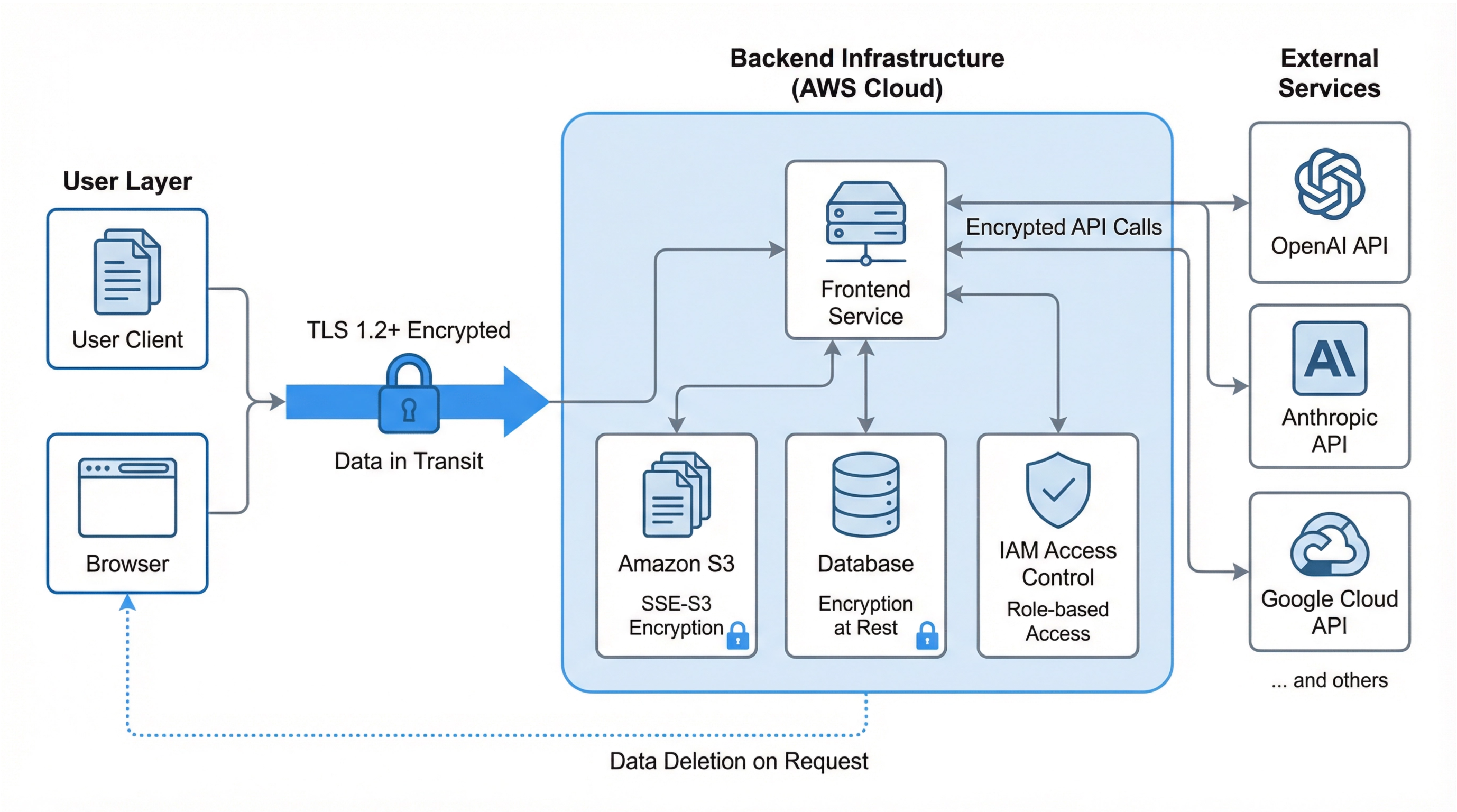}
    \caption{Security and privacy architecture of EZCollegeApp...}
    \label{fig:security-architecture}
\end{figure}

Given the sensitive nature of college application data, which includes personally identifiable information (PII), academic records, and personal essays, ensuring robust security and privacy is a paramount design principle \cite{Cavoukian2010}. The system is architected with a multi-layered security model that protects user data throughout its lifecycle, from initial upload to final deletion.

First, all data transmission between the user's browser, the EZCollegeApp front-end, and back-end services is encrypted using industry-standard Transport Layer Security (TLS) 1.2 or higher \cite{RFC8446}. This ensures that all user data, including uploaded documents and profile information, is protected against eavesdropping or man-in-the-middle attacks during transit.

Second, data at rest is securely stored using Amazon Web Services (AWS) infrastructure. Uploaded documents are stored in Amazon S3 buckets with server-side encryption (SSE-S3) enabled, which encrypts each object with a unique key. Structured data and user metadata are stored in a managed database service (e.g., Amazon DynamoDB, PostgreSQL, or RDS) with encryption at rest enabled \cite{AWSWellArchitected}. Access to these storage systems is strictly controlled through AWS Identity and Access Management (IAM) roles and policies, ensuring that only authorized back-end services can access the data.

Third, all interactions with third-party large language model providers (e.g., OpenAI, Anthropic, Google Cloud) are conducted through their secure, encrypted APIs. These providers have established data privacy protocols, which typically stipulate that data submitted via their APIs is not used for model training or stored beyond a short retention period for abuse monitoring \cite{OpenAIDataPolicy}. By leveraging these enterprise-grade APIs, the system avoids exposing sensitive user data to public or less secure model endpoints. All API keys and credentials are securely managed using AWS Secrets Manager and are never exposed on the client-side.

Finally, the system is designed to minimize the storage of sensitive data wherever possible \cite{GDPR2016}. Users retain control over their data and can request its deletion at any time, which triggers a process to permanently remove all associated documents and profile information from the system's databases and storage buckets. This comprehensive approach to security and privacy ensures that user data is handled responsibly and in accordance with best practices for sensitive information management.

\subsection{Policy Compliance by Design: Institutional AI Usage Constraints}
\label{sec:policy-compliance}

Many institutions have published guidance on the acceptable use of AI tools during the application process. While AI assistance for \textit{organizing factual information} and \textit{form filling} is often permitted, generating or substantially rewriting personal statements and essays can violate academic integrity expectations. Motivated by these constraints, EZCollegeApp adopts an \textit{extraction-over-generation} design: the system prioritizes retrieving and structuring verifiable information from official sources (e.g., deadlines, requirements, fees, contact details) instead of producing applicant-authored narrative content.

To operationalize compliance, we introduce a \textit{policy-aware interaction mode}:
\begin{itemize}
    \item \textbf{Allowed assistance:} summarizing requirements, extracting deadlines, compiling checklists, and explaining form fields.
    \item \textbf{Restricted assistance:} drafting, rewriting, or stylistically transforming essays and personal statements.
    \item \textbf{Disclosure prompts (optional):} when users request high-risk actions, the system suggests reviewing institutional policies and encourages transparent disclosure where required.
\end{itemize}

This alignment ensures EZCollegeApp remains a supportive tool for information management while reducing the risk of policy violations during admissions.

\section{Related Works}
\label{sec:related-works}

EZCollegeApp sits at the intersection of (i) AI-assisted writing and editing, (ii) domain-specific
guided workflows for high-stakes form completion, and (iii) browser-native copilots that reduce
repetitive data entry across heterogeneous portals. Below, we summarize representative commercial systems and the underlying interaction patterns that motivate our design patterns.

\subsection{AI Assistants for Writing and Communication}

A prominent class of related tools focuses on improving written communication through in-context
editing, rewriting, and drafting assistance. Grammarly is one of the best-known examples, positioning itself as an "AI assistant for communication and productivity'' that operates across a large set of applications and websites and supports workflows such as brainstorming, composing, and enhancing business communication \cite{grammarly_productivity_shift_report}.
These systems demonstrate the practical value of \emph{workflow-embedded} assistance: suggestions
appear where users write, enabling rapid iteration and immediate quality improvement.

However, writing assistants typically optimize for linguistic quality (grammar, tone, clarity) and
document-level transformation, rather than for \emph{policy-grounded correctness} under institution-specific rules. In contrast, college applications require filling structured fields and responding to prompts that are often conditional, ambiguous, and constrained by admissions policies that may change over time. This difference motivates EZCollegeApp's emphasis on (a) grounding in authoritative admissions sources, and (b) a mapping-first layer that stabilizes semantics across portals.

\subsection{Guided Expert Workflows for High-Stakes Form Completion}

Another relevant category is domain-specific "interview-style'' systems that translate complex forms into guided question-answer flows \cite{BargasAvila2011}. TurboTax is a canonical example in consumer tax preparation, providing step-by-step prompts and integrations that can reduce the burden of translating user data into formal filings \cite{Yang2018HotpotQA}. The tax domain illustrates a mature pattern for high-stakes user assistance: the user is guided through a structured workflow, and the system performs behind-the-scenes mapping into required form representations.

Relative to taxes, college admissions differ in two key ways. First, many application questions are
open-ended and narrative (e.g., activity descriptions and short essays), where correctness involves both faithfulness to personal materials and appropriate presentation. Second, requirements and interpretations are distributed across institution-specific webpages and FAQs, making retrieval and provenance critical.

Accordingly, EZCollegeApp borrows the \emph{guided-workflow} intuition but emphasizes transparent,
cited evidence and explicit human control, rather than opaque end-to-end automation \cite{Amershi2019}.

\subsection{Career and Job-Application Copilots}

A third closely related line of systems focuses on job discovery and application submission across
fragmented applicant tracking systems (ATS) and company portals. Simplify presents itself as a "common application for jobs \& internships" and its Copilot browser extension supports one-click auto-fill for applications, resume tailoring, and automatic application tracking.

This product family is especially relevant because the job-application workflow mirrors college
applications in its \emph{cross-portal heterogeneity}: field labels differ across sites, forms contain dynamic
sections, and applicants repeatedly re-enter the same underlying profile.

At the same time, job-application tools vary widely in how they trade off convenience and control.
Some offerings emphasize aggressive automation, including form-filling and even end-to-end submission on supported platforms. In high-stakes educational contexts, however, automated
submission and opaque transformations can create unacceptable risks (e.g., misrepresentation, policy non-compliance, or accidental disclosure). EZCollegeApp, therefore, adopts a strict human-in-the-loop pattern: the system presents suggestions with citations, but users remain the final authors and operators of submission actions.

\subsection{Admissions Platforms and Planning Tools}

Beyond AI copilots, students commonly rely on centralized application portals and planning tools to manage tasks, deadlines, and school-specific requirements. While these platforms reduce logistical friction, they generally do not provide (i) field-level semantic normalization across heterogeneous portals, (ii) evidence-grounded answers linked to authoritative admissions documents, or (iii) a unified, user-owned knowledge base built from both personal materials (transcripts, resumes, essays, activities, honors, etc) and institutional policy.

EZCollegeApp targets these gaps by combining a canonical schema and mapping engine with retrieval-augmented generation and explicit provenance tracking.

\subsection{Applications of Large Language Models Across Domains}
\label{subsec:llm-applications}

LLMs have rapidly evolved from general-purpose text generators into reusable foundation models that can be adapted to a wide range of domain workflows. A prevailing pattern in recent systems is to combine LLMs with (i) external knowledge access (e.g., RAG), (ii) tool use and programmatic execution (e.g., calling calculators, databases, or coding assistants), (iii) structured intermediate representations, and (iv) multimodal inputs and outputs (documents, images, tables). These techniques enable LLMs to operate in settings that demand higher factuality, stronger constraint satisfaction, and clearer provenance than unconstrained free-form generation.

Consequently, recent research has investigated LLM-based tools across numerous fields, such as industrial production, genomics, teaching, clinical diagnostics and cancer therapy, banking, drug development, software engineering and automation, quantum technology, farming, connected devices, neuroscience, computational physics, creative disciplines, underrepresented linguistic communities, virtual simulations, legal practice, governance studies, and textual analysis~\cite{tan2023promises,liu2023transformation,ma2023impressiongpt,liao2023differentiate,dai2023adautogpt,liu2023summary,guan2023cohortgpt,cai2022coarse,liu2023pharmacygpt,shi2023mededit,gong2023evaluating,liu2022survey,cai2023coarsetofine,liao2023maskguided,rezayi2023exploring,liu2023radiology,ruan2026large,liu2025survey,liu2025pharmacygpt,gong2024advancing,liu2025aimanities,fang2026knowledgedistillationdatasetdistillation,dou2023towards,shankar2025bridging,pan2025bridging,shu2024transcending,zhong2024opportunities,liu2023radonc,zhou2026digital,liu2024llm,wang2024legal,yang2024analyzing,li2024large,shu2025momoe,most2024evaluating,wu2025exploring,wang2025prompt,zhao2025urban,zhao2024revolutionizing,zhang2023biomedgpt,zhao2023brain,liao2023differentiating,huang2024position,wang2024comprehensive,li2024artificial,liu2023context,dou2023towards,rezayi2024exploring,liu2023holistic,tan2023promises,guan2023cohortgpt,liu2023pharmacygpt,zhong2025chatabl,liu2024radiation,liu2024fine,liu2025ad}.

EZCollegeApp aligns with these cross-domain trends but targets a distinct high-stakes workflow: heterogeneous, conditional, and policy-constrained college application forms. In this context, helpfulness is inseparable from correctness and accountability. We therefore adopt a mapping-first abstraction to stabilize semantics across portals, and evidence-grounded answer synthesis with explicit provenance, enabling users to verify and override suggestions. This design positions EZCollegeApp as a domain workflow copilot that prioritizes transparency and human control—properties that are increasingly central across LLM deployments in other regulated or high-consequence domains.

\subsection{Summary}
Taken together, these systems motivate three design lessons adopted by EZCollegeApp with workflow-embedded assistance; structured mapping between the userinputs and formal forms existed in guided tax workflows; and cross-portal field normalization and reusable profiles sourced from job-application copilots.  
Then, EZCollegeApp extends these patterns with a mapping-first architecture and source-grounded answer synthesis to meet the correctness and accountability requirements of college admissions. Finally, consistent with broader cross-domain LLM deployments, EZCollegeApp emphasizes domain grounding, constraint-aware generation, and explicit provenance to support high-stakes decision contexts.

% ======================================================
% ======================================================
\section{System Overview}
\label{sec:system-overview}

EZCollegeApp is implemented as a distributed, cloud-native system composed of multiple specialized components that collectively assist students throughout the American college application lifecycle \cite{Burns2016}. The system is designed around a strict separation of concerns, enabling independent evolution of user interaction, document understanding, reasoning, and deployment infrastructure \cite{Saltzer1984}. Rather than operating as a single monolithic application, EZCollegeApp adopts a modular architecture consisting of three loosely coupled subsystems: a conversational front-end for user orchestration, a back-end reasoning and document processing platform, and a browser-based Application Copilot that delivers real-time, human-supervised assistance during form completion \cite{Parnas1972}.

All components communicate through authenticated, versioned RESTful APIs \cite{Fielding2000}. User authentication and authorization are handled using JSON Web Tokens (JWT), ensuring that all data access is auditable and scoped to individual users. This architectural design prioritizes scalability, robustness, and maintainability while preserving strong guarantees around user control and data provenance \cite{Sculley2015}.

\subsection{High-Level Architecture}

The back-end infrastructure functions as the central coordination layer of EZCollegeApp, managing data persistence, reasoning workflows, and communication between user-facing interfaces. The major architectural components are summarized below.

\begin{enumerate}[label=\arabic*), wide, labelindent=0pt]

\item \textbf{Conversational Front-End}

The primary user entry point is a web-based conversational interface implemented using Next.js and TypeScript. This component serves as an intelligent orchestration layer rather than a passive upload portal. Through a guided, multi-turn dialogue, the interface collects user intent (e.g., target institutions, application timelines) and prompts the user to upload relevant source materials, including academic transcripts, curriculum vitae (CV), activity lists, and personal essays.

This interaction model reduces onboarding friction and ensures that a complete and coherent set of documents is collected prior to downstream processing.

\item \textbf{Document Processing and Reasoning Pipeline}

The back-end processing layer is implemented as a hybrid system leveraging Go for high-throughput services and Python (FastAPI) for AI-driven document understanding. Once documents are uploaded, an asynchronous, multi-stage pipeline is triggered:

\begin{itemize}
    \item \textbf{Ingestion and Storage}: Uploaded files are securely stored in an Amazon S3 bucket. File-level metadata (e.g., user ID, document type, upload timestamp) is recorded in a DynamoDB table, decoupling raw file storage from processing logic.
    
    \item \textbf{Document Classification and Extraction}: A Python-based service performs initial document classification to identify file types (e.g., transcript, CV, essay). Text extraction is performed using \path{pdfplumber} for digital PDFs and \path{pytesseract} for OCR on scanned documents. The extracted text is then passed to a large language model, which performs structured information extraction into a normalized JSON representation covering multiple semantic categories (e.g., education history, grades, awards, activities).
    
    \item \textbf{Semantic Chunking and Indexing}: Extracted content is segmented into semantically coherent chunks. Each chunk is embedded and indexed in an Amazon OpenSearch cluster, together with metadata such as user ID, source document, and content category. This process constructs a per-user knowledge base optimized for retrieval-augmented generation (RAG).
\end{itemize}

\item \textbf{Application Copilot (Browser Extension)}

The Application Copilot is implemented as a Manifest V3 Chrome extension written in JavaScript. It serves as the system's human-in-the-loop interface during application form completion. After authenticating with the back-end, the extension retrieves and locally caches the user's structured profile data.

When the user navigates to a supported college application portal, the extension's content scripts activate and perform real-time analysis of the page's DOM. Form elements and associated labels are identified, mapped to internal canonical fields, and matched with relevant profile data. Suggested content is presented through a lightweight, non-intrusive overlay. Crucially, the extension never automatically fills or submits form fields; all content transfer requires explicit user action.

\end{enumerate}

This service-oriented architecture, deployed on scalable cloud infrastructure, enables fault tolerance, independent scaling of subsystems, and continuous system evolution without compromising user trust or control.

\subsection{Workflow Summary}

The end-to-end user workflow in EZCollegeApp is designed as a progressive transformation from unstructured personal documents to contextual, human-supervised assistance \cite{Manning2008}.

First, the user interacts with the conversational front-end, which guides document upload and intent specification. Uploaded materials trigger the asynchronous document processing pipeline, which classifies, extracts, structures, and indexes user information \cite{Kleppmann2017}. The result is a canonical, machine-readable representation of the student profile, backed by a searchable knowledge base.

When the user later accesses a college application website, the Application Copilot activates automatically. As the user focuses on individual form fields, the copilot identifies the semantic intent of each field, retrieves relevant profile information, and presents one or more candidate suggestions \cite{Yu2018Spider}. The user may choose to copy suggested content to the clipboard and manually paste it into the form. This deliberate suggest--copy--paste interaction enforces human oversight and preserves applicant agency at all times.

\begin{figure}[htbp]
    \centering
    \includegraphics[width=\textwidth]{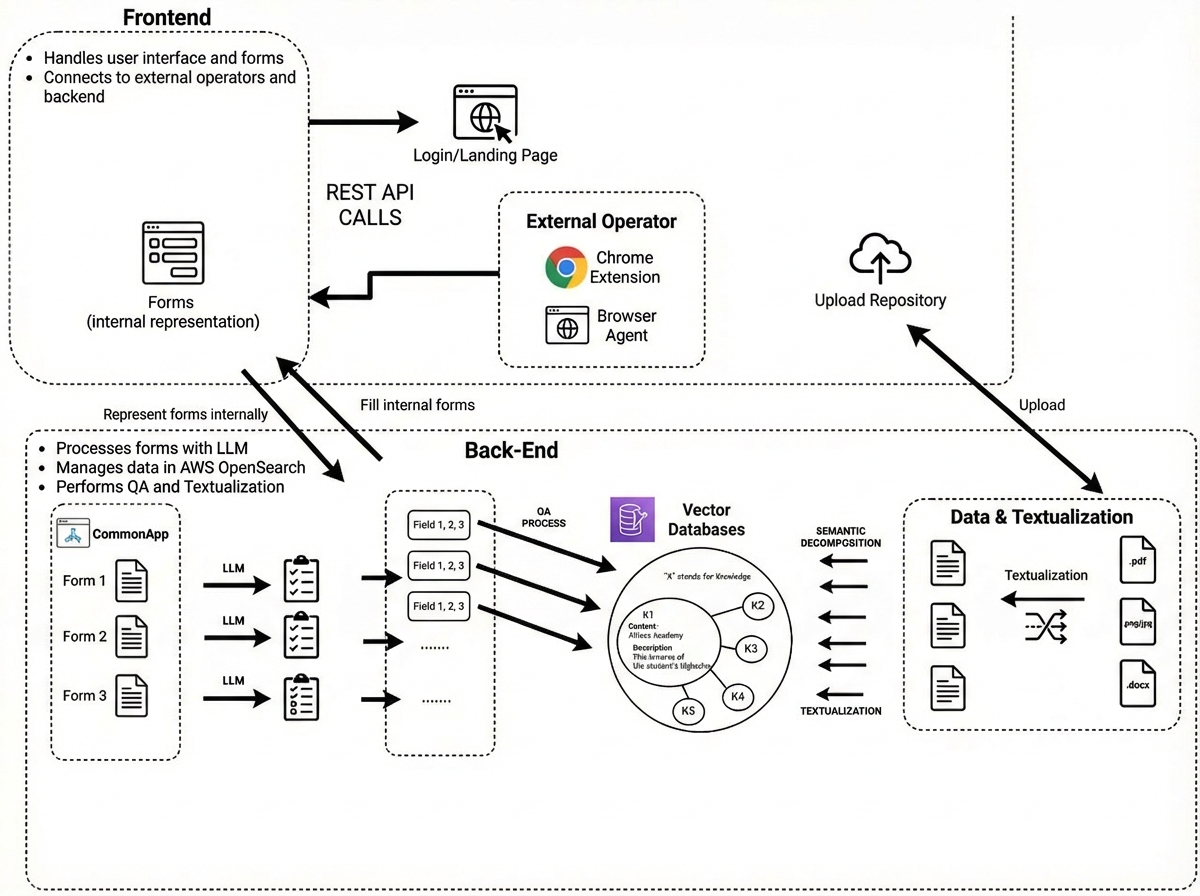}
    \caption{System architecture and workflow of EZCollegeApp. The system consists of three main components: (1) a conversational front-end for document collection and user orchestration, (2) a back-end document processing and reasoning pipeline that extracts, structures, and indexes student information, and (3) an Application Copilot browser extension that provides real-time, human-supervised assistance during form completion. The workflow progresses from document upload through processing and indexing to contextual answer suggestion.}
    \label{fig:system-overview}
\end{figure}

% ======================================================
\section{Information Collection and Processing}
\label{sec:data-pipeline}

The construction of a reliable and comprehensive knowledge base for college admissions requires systematic collection, processing, and indexing of information from diverse sources. EZCollegeApp implements a multi-stage data pipeline that transforms unstructured web content into a structured, searchable corpus optimized for retrieval-augmented generation. This pipeline is designed to handle the scale and heterogeneity of admissions information across hundreds of institutions while maintaining data quality, provenance tracking, and incremental updateability.

\subsection{Admissions Website Crawling}

Admissions webpages for each institution are systematically crawled using a custom-built web scraping infrastructure implemented with the Playwright browser automation framework. The crawling process is designed to be both comprehensive and respectful of institutional server resources, balancing coverage with politeness.

\subsubsection{Crawl Initialization and Target Selection}

The crawling process begins with a curated CSV file containing the official website URLs for target institutions, typically sourced from authoritative directories such as the U.S. Department of Education's College Scorecard or the Common Application member list. Each institution is identified by its root admissions URL (e.g., \path{https://admission.stanford.edu} ), which serves as the entry point for the crawler.

\subsubsection{Keyword-Guided Breadth-First Search}

For each institution, the crawler employs a breadth-first search (BFS) strategy augmented with keyword-based prioritization to focus on admissions-relevant content. The system maintains two priority queues: a high-priority queue for pages whose URLs or anchor text contain admissions-related keywords (e.g., "application requirements," "deadlines," "testing policies," "financial aid," "international applicants," "transcript request," "letters of recommendation"), and a low-priority queue for other pages within the same domain. This prioritization ensures that the most relevant pages are crawled first, even when the crawl budget is limited.

The crawler extracts all hyperlinks from each visited page using BeautifulSoup, normalizes them to absolute URLs, and filters them based on domain matching to ensure that only pages within the institution's official domain are queued. Links to non-HTML resources (e.g., PDFs, images, videos) and external domains are excluded from the crawl queue, though external resource URLs are logged for potential future analysis.

\subsubsection{Dynamic Content Rendering and Dual-Format Storage}

Modern admissions websites frequently employ JavaScript-heavy frameworks that render content dynamically, making traditional static HTML scraping insufficient. To address this, the crawler uses Playwright's headless Chromium browser to fully render each page, executing JavaScript and waiting for the DOM to stabilize (via \path{wait\_until="domcontentloaded"} and explicit \path{wait\_for\_selector("body")} calls). This ensures that dynamically loaded content, such as accordion menus, tabbed interfaces, and lazy-loaded sections, is captured.

Each rendered page is saved in two formats:
\begin{itemize}
    \item \textbf{HTML}: The raw HTML source is saved for text extraction and link discovery. This format preserves the document structure and metadata (e.g., \path{<meta>} tags, \path{<title>}).
    \item \textbf{PDF}: The page is rendered to a PDF using Playwright's \path{page.pdf()} method with A4 formatting. This format provides a visual snapshot of the page as it appeared at crawl time, preserving layout, images, and styling. PDFs serve as an archival format for human review and as a fallback for text extraction when HTML parsing fails.
\end{itemize}

This dual-format storage strategy balances machine readability (HTML) with human readability (PDF), and provides redundancy in case one format becomes corrupted or inaccessible.

\subsubsection{Politeness and Rate Limiting}

To avoid overloading institutional servers and to minimize the risk of being blocked by anti-bot mechanisms, the crawler implements several politeness measures:
\begin{itemize}
    \item \textbf{Randomized Delays}: A random delay of 8-15 seconds is inserted between consecutive requests to the same domain. This delay range is chosen to mimic human browsing behavior while maintaining reasonable throughput.
    \item \textbf{Crawl Budget}: Each institution is subject to a maximum page limit (typically 50 pages), preventing the crawler from exhausting server resources on large sites with extensive non-admissions content.
    \item \textbf{Parallel Crawling with Concurrency Control}: Multiple institutions are crawled in parallel using an async/await architecture with semaphore-based concurrency control (typically 3-5 concurrent institutions). This parallelism is applied at the institution level, not the page level, to avoid overwhelming any single server.
    \item \textbf{Resume Capability}: The crawler maintains a checkpoint file for each institution, recording which pages have been successfully crawled. If the crawl is interrupted (e.g., due to network errors or rate limiting), it can resume from the last checkpoint without re-crawling already-processed pages.
\end{itemize}

\subsubsection{Boilerplate Removal}

Admissions webpages typically contain substantial boilerplate content that is repeated across multiple pages, such as navigation menus, site headers and footers, cookie consent banners, and social media links. Including this content in the knowledge base would introduce noise and dilute the relevance of retrieval results. To address this, the system employs a combination of rule-based and heuristic methods for boilerplate detection and removal:

\begin{itemize}
    \item \textbf{DOM-Based Filtering}: During HTML parsing, elements with specific CSS classes or IDs commonly associated with boilerplate (e.g., \path{class="nav"}, \path{id="footer"}, \path{class="cookie-banner"}) are excluded from text extraction.
    \item \textbf{Repetition Analysis}: Text blocks that appear verbatim on multiple pages from the same institution are flagged as potential boilerplate. Blocks that exceed a repetition threshold (e.g., appearing on more than 50\% of pages) are removed.
    \item \textbf{Content Density Heuristics}: Regions of the page with low text density (high ratio of HTML tags to text) or high link density (many hyperlinks per word) are deprioritized or excluded, as they are likely to be navigation elements rather than substantive content.
\end{itemize}

\subsubsection{Layout-Aware Parsing}

To preserve the semantic structure of admissions documents, the text extraction process employs layout-aware heuristics that identify and retain section boundaries, headings, and list structures. HTML heading tags (\path{<h1>} through \path{<h6>}) are mapped to hierarchical section markers, and list elements (\path{<ul>}, \path{<ol>}, \path{<li>}) are preserved with appropriate indentation or bullet points. This structural information is encoded as lightweight markup (e.g., Markdown-style \path{\#\#} for headings) in the extracted text, enabling downstream chunking algorithms to respect topical boundaries.

\subsubsection{Error Handling and Logging}

The crawler is designed to be resilient to transient failures. Network timeouts, HTTP errors (e.g., 404, 500), and JavaScript execution errors are caught and logged, but do not halt the overall crawl. Failed pages are retried up to three times with exponential backoff before being marked as permanently failed. All errors are logged to a separate error log file with timestamps, URLs, and error messages, enabling post-crawl analysis and manual review of problematic pages.

\subsubsection{Output Organization}

For each institution, the crawler creates a directory structure organized by domain name (sanitized to remove special characters). Within each institution's directory, a \path{pages/} subdirectory contains all HTML and PDF files, named according to a sanitized version of the page URL. An \path{index.json} file at the root of each institution's directory provides metadata for all crawled pages, including the URL, page title, file paths for HTML and PDF, and crawl timestamp. This structured output facilitates downstream processing and enables efficient lookup of source documents during answer generation.

\subsubsection{Incremental Updates and Versioning}

Admissions policies and requirements change frequently, necessitating periodic re-crawling and re-indexing of institutional websites. To support incremental updates without disrupting the live system, the pipeline implements a versioning strategy:

\begin{itemize}
    \item \textbf{Crawl Versioning}: Each crawl is assigned a unique version identifier (e.g., a timestamp or sequential version number). Newly crawled pages are stored in a separate directory or database partition, allowing the old version to remain accessible until the new version is validated.
    \item \textbf{Differential Indexing}: When re-crawling an institution, the system compares the new page content with the previous version (using content hashing or text similarity). Only pages that have changed are re-processed and re-indexed, reducing computational overhead.
    \item \textbf{Graceful Rollover}: Once the new version is fully indexed and validated, the system atomically switches the active index to the new version, ensuring zero downtime for end users.
\end{itemize}

This versioning strategy enables the system to maintain up-to-date information while preserving historical snapshots for auditing and analysis.

\subsection{Text Extraction and Cleaning}

Once admissions webpages have been crawled and stored, the next stage of the pipeline extracts machine-readable text from the saved documents and removes extraneous content that would degrade retrieval quality.

\subsubsection{Document and Image Text Extraction}

Text extraction from PDF files and images is performed using a multi-stage pipeline that combines rule-based extraction, optical character recognition, and vision-language models to handle diverse document formats robustly \cite{Smith2007}.

For PDF files, the system first attempts extraction using the \path{pdfplumber} library, which provides robust handling of native text PDFs where text is encoded as selectable characters \cite{Manning2008}. The library preserves spatial layout information, such as column boundaries and reading order, enabling accurate reconstruction of document structure. However, if the extracted text is too sparse (fewer than 50 characters per page) or if extraction fails entirely, the system falls back to optical character recognition using \path{pytesseract}, which invokes the Tesseract OCR engine to recognize text from rendered page images. This fallback mechanism ensures reliable text extraction even for scanned documents or PDFs with embedded images rather than encoded text.

For image files (such as PNG, JPEG, or TIFF formats), the system employs a vision-language model, typically GPT-4o, to extract textual content \cite{Radford2021, Li2023BLIP2}. The image is encoded and sent to the model with a prompt instructing it to transcribe all visible text while preserving layout and formatting where possible \cite{Xu2020LayoutLM}. This approach is particularly effective for screenshots of web pages, photographs of handwritten documents, or images containing mixed text and graphical elements. The vision-language model can also interpret context and structure in ways that traditional OCR systems cannot, such as recognizing tables, captions, and hierarchical document organization.

The extracted text from all sources undergoes several normalization steps to ensure consistency and compatibility with downstream processing:
\begin{itemize}
    \item \textbf{Whitespace Normalization}: Multiple consecutive spaces, tabs, and newlines are collapsed into single spaces or line breaks to reduce redundancy.
    \item \textbf{Unicode Normalization}: Text is normalized to the NFC (Canonical Decomposition followed by Canonical Composition) form to ensure consistent representation of accented characters and special symbols.
    \item \textbf{Encoding Error Correction}: Common encoding artifacts (such as mojibake from UTF-8 and Latin-1 mismatches) are detected and corrected using heuristic rules.
\end{itemize}

\subsection{Semantic Chunking and Indexing}

The cleaned and structured text is then segmented into semantically coherent chunks suitable for embedding and retrieval. Chunking is a critical step in the pipeline, as chunks that are too small may lack sufficient context for accurate retrieval, while chunks that are too large may conflate multiple topics and reduce retrieval precision.

\subsubsection{Chunking Strategy}

EZCollegeApp employs a hybrid chunking strategy that balances fixed-size constraints with semantic boundaries:

\begin{itemize}
    \item \textbf{Target Chunk Size}: Each chunk is targeted to contain 300-500 tokens (approximately 200-350 words), a range empirically determined to provide sufficient context for embedding models while remaining concise enough for LLM context windows.
    \item \textbf{Boundary Respect}: Chunks are not permitted to split in the middle of a sentence or paragraph. The chunking algorithm identifies natural breakpoints (e.g., paragraph boundaries, section headers) and extends or contracts the chunk size slightly to align with these boundaries.
    \item \textbf{Overlap}: Adjacent chunks include a small overlap (typically 50 tokens) to ensure that information spanning a chunk boundary is not lost. This overlap also improves retrieval recall by providing multiple entry points for queries that match content near a boundary.
\end{itemize}

The chunking algorithm processes the cleaned text sequentially, accumulating tokens until the target size is reached, then backtracking to the nearest paragraph or section boundary to finalize the chunk. This approach ensures that each chunk is a self-contained, coherent unit of information.

\subsubsection{Metadata Annotation}

Each chunk is annotated with rich metadata to support filtering, ranking, and provenance tracking during retrieval:

\begin{itemize}
    \item \textbf{Institution}: The name of the university or college from which the content was sourced.
    \item \textbf{Source URL}: The original webpage URL from which the chunk was extracted.
    \item \textbf{Page Title}: The title of the source webpage, providing additional context for the chunk's topic.
    \item \textbf{Section Heading}: If the chunk falls under a specific section heading (e.g., "Application Deadlines," "Testing Requirements"), that heading is recorded as metadata.
    \item \textbf{Source Type}: A categorical label indicating whether the chunk originates from an official admissions website, a curated FAQ, or community content.
    \item \textbf{Crawl Timestamp}: The date and time when the source page was crawled, enabling recency-based ranking and identification of stale content.
    \item \textbf{Chunk Index}: The position of the chunk within the source document, allowing reconstruction of the original document order if needed.
\end{itemize}

\subsection{Indexing and Retrieval Architecture}

The system supports flexible indexing strategies to accommodate different deployment scenarios and performance requirements. The architecture is designed to enable seamless switching between retrieval methods based on computational resources, deployment environment, and query characteristics.

\subsubsection{Embedding and Vector Database Indexing}

Each semantically chunked passage is encoded into a dense vector representation using a sentence transformer model (e.g., \path{all-MiniLM-L6-v2} or \path{sentence-transformers/all-mpnet-base-v2}). These models are pre-trained on large corpora and fine-tuned for semantic similarity tasks, producing embeddings that capture the meaning of the text in a high-dimensional space (typically 384 or 768 dimensions).

The embeddings, along with their associated metadata, are stored in a vector database that supports efficient approximate nearest neighbor (ANN) search. EZCollegeApp supports two vector database backends:
\begin{itemize}
    \item \textbf{ChromaDB}: A lightweight, open-source vector database optimized for local deployment and rapid prototyping. ChromaDB (\cite{Chromacorechroma}) stores embeddings and metadata in a local directory and provides a simple Python API for indexing and querying.
    \item \textbf{Amazon OpenSearch}: A scalable, cloud-based search and analytics engine with native support for vector search via the k-NN plugin. OpenSearch (\cite{amazonOpenSearch}) is used for production deployments requiring high availability, horizontal scalability, and integration with other AWS services.
\end{itemize}

During indexing, each chunk is assigned a unique document ID (e.g., \texttt{doc\_<institution>\_<page\_id>\_\allowbreak<chunk\_index>} and stored with its embedding vector and metadata. The vector database builds an ANN index (such as HNSW or IVF) to enable sub-linear time retrieval of the top-k most similar chunks for a given form field. All chunks are stored with rich metadata, including source institution, URL, page title, timestamp, and source type, enabling consistent filtering, ranking, and provenance tracking.

\subsubsection{Vector-Based Retrieval}

When the system encounters a form field that requires information from user documents, the form question is embedded using the same sentence transformer model employed during indexing \cite{Reimers2019}. The system then retrieves the top-k most similar chunks from the vector database using cosine similarity or approximate nearest neighbor search \cite{Karpukhin2020}. This approach excels at capturing semantic similarity and handling questions that may be phrased differently across various application portals, making it particularly effective when different institutions ask for the same information using different terminology \cite{Guo2016}.

Vector-based retrieval can identify relevant passages even when there is no direct keyword overlap between the form question and the document text. For example, a form field asking about "extracurricular leadership positions" can successfully retrieve passages mentioning "club president" or "team captain" based on semantic similarity. 

% \subsubsection{Vectorless Retrieval}

% Alternatively, the system can employ vectorless, reasoning-based retrieval methods such as PageIndex (\cite{PageIndex}), which organize documents hierarchically by structure (such as section headings, page boundaries, and document organization) and use keyword-based matching combined with contextual reasoning to identify relevant passages. This approach eliminates the need for embedding computation and vector storage, significantly reducing infrastructure costs, memory requirements, and query latency.

% Vectorless methods are particularly effective for structured documents with clear organizational hierarchies, such as resumes with well-defined sections like "Education," "Experience," and "Awards." They also perform well when form questions align closely with document section headings and terminology, such as when a form field requests "GPA" and the user's transcript explicitly contains this label. The system can dynamically select between vector-based and vectorless retrieval based on the form field type, document structure, and available computational resources, ensuring optimal performance across diverse deployment scenarios.

\subsubsection{Structure-Aware Retrieval}
\label{subsubsec:structure-aware-retrieval}

As an alternative to vector-based methods, EZCollegeApp employs structure-aware retrieval that leverages the inherent hierarchical organization of student documents \cite{bast2013index}. Upon ingestion, the system uses multimodal language models to construct a tree representation of each document, identifying section boundaries through table-of-contents detection or structural inference from heading patterns \cite{Kiyono2020}. Each node is annotated with a unique identifier, title, page range, and a concise summary. At query time, the LLM evaluates node titles and summaries to select relevant sections, then extracts content using a page-to-text mapping \cite{Yao2023ReAct}. This approach eliminates embedding computation and vector database dependencies while naturally preserving document structure and enabling precise source attribution with section titles and page numbers \cite{Rashkin2021}.

Structure-aware retrieval is particularly effective for student application materials with clear organizational hierarchies, such as resumes with distinct ``Education,'' ``Experience,'' and ``Awards'' sections, or transcripts with well-defined course listings. The system dynamically selects between vector-based and structure-aware retrieval based on document characteristics and query type \cite{Lin2021}.

\subsubsection{Storage and Scalability Considerations}

Due to the scale and size of the corpus—comprising tens of thousands of webpages across hundreds of institutions—the raw HTML and PDF files are stored locally on dedicated storage servers rather than on cloud drives \cite{Hennessy2019}. This local storage approach reduces cloud storage costs and egress fees, while still providing sufficient redundancy through RAID configurations and periodic backups to cloud object storage (e.g., Amazon S3) for disaster recovery \cite{McSherry2015}.

The vector embeddings and metadata, being more compact and frequently accessed, are stored in the vector database (ChromaDB or OpenSearch), which is optimized for low-latency retrieval \cite{Johnson2019}. This hybrid storage architecture balances cost, performance, and scalability, enabling the system to handle a growing corpus without prohibitive infrastructure expenses.

% ======================================================
% ======================================================
\section{Internal Form Representation and Mapping}
\label{sec:form-mapping}

A core technical challenge in building a universal college application assistant lies in the extreme heterogeneity of application platforms. Identical concepts may appear under different labels, structures, and conditional logic across portals such as the Common Application, the Coalition Application, and institution-specific systems. EZCollegeApp addresses this challenge through an abstraction layer built around a canonical form schema and a multi-tiered mapping engine.

\subsection{Canonical Form Schema}

At the foundation of the system is a canonical form schema that serves as a universal ontology for U.S. college application data. The schema is defined as a version-controlled JSON specification and captures both semantic meaning and structural constraints for each field. Each schema entry includes:

\begin{itemize}
\item \textbf{Field Identifier}: A unique hierarchical identifier (e.g., \path{user.profile.education.high\_school.gpa}).
\item \textbf{Semantic Intent}: A concise description of the field's conceptual meaning, used for disambiguation and semantic matching.
\item \textbf{Data Type}: A strict type specification (e.g., String, Number, Date, Boolean, Enum) that informs validation and UI expectations.
\item \textbf{Keywords and Synonyms}: A comprehensive list of alternative labels and phrasings commonly used in application forms.
\item \textbf{Formatting Rules}: Constraints specifying expected input formats (e.g., date formats, numeric precision).
\item \textbf{Conditional Logic}: Explicit dependency rules that define field visibility based on other fields.
\item \textbf{Provenance}: Metadata tracking the origin of extracted data (e.g., transcript vs.\ CV), enabling conflict resolution and confidence assessment.
\end{itemize}

This schema can be extended or updated without modifying core system logic, allowing the platform to adapt to new forms and evolving requirements.

\subsection{Mapping External Forms to Internal Fields}

The Application Copilot maps HTML form elements to canonical schema fields using a resilient, real-time, three-tier mapping engine executed entirely within the browser:

\begin{enumerate}[label=\arabic*), wide, labelindent=0pt]

\item \textbf{Tier 1: Direct Semantic Matching}

The engine aggregates all text associated with a form field, including linked \path{<label>} elements, placeholders, ARIA labels, and nearby text nodes. After normalization, this text is matched against the schema's keyword and synonym lists using deterministic string matching. This tier resolves the majority of well-labeled fields with minimal computational cost.

\item \textbf{Tier 2: Contextual and Structural Analysis}

If direct matching is inconclusive, the engine examines the surrounding DOM structure. Parent containers, element IDs, class names, and grouping semantics are analyzed to infer contextual meaning. For example, a generic 'City' field can be correctly mapped when embedded within an address-related container. Accessibility metadata is also leveraged to improve robustness.

\item \textbf{Tier 3: Semantic Similarity Scoring}

For highly ambiguous fields, the engine computes semantic similarity scores between the aggregated field text and the schema's semantic intent descriptions. Lexical similarity metrics (e.g., Jaro--Winkler distance, word overlap coefficients) are combined into a confidence score. Mappings are accepted only when the score exceeds a predefined threshold, reducing false positives.

\end{enumerate}

Mapping results are cached in session storage to avoid redundant computation during repeated interactions.

\subsection{Conditional Logic and Dependencies}

Dynamic forms frequently reveal or hide fields based on prior user input. To ensure relevance, the Application Copilot enforces the conditional logic encoded in the canonical schema. The extension registers event listeners on dependency fields and maintains an internal dependency graph. When a user modifies a controlling field, the rules engine updates the field's visibility state in real time, ensuring that suggestions are generated only for currently active fields. This dynamic awareness enables accurate, context-sensitive assistance without user disruption.

% ======================================================
\section{Question Answering for Form Completion}
\label{sec:qa-forms}

The core functionality of EZCollegeApp is to assist students in completing college application forms by automatically generating contextually appropriate answers based on their uploaded documents and profile information. Unlike traditional form-filling systems that rely on simple template matching or direct field-to-database lookups, EZCollegeApp employs a sophisticated question-answering pipeline that combines semantic understanding, retrieval-augmented generation, and constraint-aware synthesis. This approach ensures that generated answers are not only factually grounded in the student's actual experiences but also conform to the structural and formatting requirements of each application form.

\subsection{Mapping-First Answer Generation}

EZCollegeApp follows a mapping-first approach to answer generation, which decouples the understanding of form questions from the synthesis of answers. Rather than generating answers directly from raw form questions, the system first resolves each question to a canonical field in the internal schema and then formulates a structured query based on the field's semantic intent and data type. This two-stage process provides several advantages over direct question-to-answer generation.

\subsubsection{Advantages of the Mapping-First Paradigm}

The mapping-first approach offers three key benefits that improve both the accuracy and consistency of generated answers:

\paragraph{Cross-Portal Consistency}

Different application portals often ask semantically equivalent questions using different wording. For example, the Common Application might ask "Please list your extracurricular activities," while a university-specific portal might ask "Describe your involvement in school and community organizations." By mapping both questions to the same canonical field (e.g., \path{user.activities.list}), the system ensures that the same underlying information is used to answer both questions, maintaining consistency across applications \cite{Rahm2001}. This is particularly important for students applying to multiple institutions, as inconsistencies in reported activities or achievements can raise red flags during admissions review.

\paragraph{Semantic Disambiguation}

Raw form questions are often ambiguous or context-dependent. For example, the question "When did you start?" could refer to the start date of an activity, the start date of high school, or the start date of employment, depending on the surrounding form context \cite{Zettlemoyer2005}. The mapping process resolves this ambiguity by analyzing the question's position in the form hierarchy, the labels of parent sections, and the data type constraints specified in the canonical schema. Once the question is mapped to a specific canonical field (e.g., \path{user.activities[0].start\_date}), the system can generate an unambiguous query that retrieves the correct information.

\paragraph{Type-Aware Query Formulation}

Each canonical field in the schema is annotated with a data type (e.g., date, number, text, multiple-choice) and formatting constraints (e.g., MM/DD/YYYY for dates, 10-digit format for phone numbers) \cite{Dziri2022}. By mapping the form question to a canonical field before generating an answer, the system can formulate a query that explicitly specifies the expected output format. This ensures that the language model's response conforms to the field's constraints, reducing the need for post-processing or error correction. For example, if a question is mapped to a date field, the query instructs the model to return the answer in MM/DD/YYYY format, and the system validates the output against a date regex before inserting it into the form.

\subsubsection{Mapping Process}

The mapping process is executed by the three-tier matching engine described in Section 4. For each form question, the system performs the following steps:

\begin{enumerate}
    \item \textbf{Tier 1 - Keyword Matching}: The system extracts keywords from the question text and compares them against the predefined keyword lists for each canonical field. If a direct match is found (e.g., the question contains "GPA" and the canonical field \path{user.academics.gpa} has "GPA" as a keyword), the question is immediately mapped to that field with high confidence.
    
    \item \textbf{Tier 2 - Contextual Analysis}: If no direct keyword match is found, the system analyzes the question's context by examining the DOM structure of the form. It considers the labels of parent sections, sibling fields, and HTML attributes (e.g., \path{name}, \path{id}, \path{placeholder}). For example, if the question "Start date" appears within a section labeled "Extracurricular Activities," the system infers that it refers to \path{user.activities[i].start\_date} rather than \path{user.education.start\_date}.
    
    \item \textbf{Tier 3 - Semantic Similarity}: If contextual analysis is insufficient, the system computes the semantic similarity between the question text and the semantic intent descriptions of all canonical fields. This is done using word overlap and Levenshtein distance metrics. The field with the highest similarity score (above a threshold of 0.85) is selected as the mapping target.
\end{enumerate}

Once a mapping is established, the system constructs a structured query that includes the canonical field identifier, the expected data type, and any relevant context from the student's profile.

\subsection{Grounded Retrieval and Answer Synthesis}

After mapping a form question to a canonical field, the system retrieves relevant document chunks from the vector database and synthesizes an answer using a retrieval-augmented generation approach. This process ensures that all generated answers are grounded in factual evidence from the student's uploaded documents, rather than being hallucinated or fabricated by the language model.

\subsubsection{Query Construction and Embedding}

The system constructs a natural language query based on the canonical field's semantic intent and the original form question \cite{Zettlemoyer2005}. For example, if the form question "What is your cumulative GPA?" is mapped to the canonical field \path{user.academics.gpa}, the system constructs a query such as "What is the student's cumulative GPA or grade point average?" This query is then processed by the retrieval system \cite{Carpineto2012}. In vector-based configurations, the query is embedded into a high-dimensional vector using a pre-trained sentence embedding model, capturing its semantic meaning for similarity-based retrieval. In vectorless configurations, the query is analyzed for keywords and semantic intent, which are used to navigate the document index structure and identify relevant passages through reasoning-based matching.

\subsubsection{Document Retrieval}

The processed query is used to retrieve the top-k most relevant document chunks from the indexed knowledge base. In vector-based retrieval, the query embedding is matched against chunk embeddings using cosine similarity as the distance metric \cite{Reimers2019}, with the system typically retrieving the top 5-10 chunks for each query \cite{Karpukhin2020}. In vectorless retrieval, the system navigates the document hierarchy using keyword matching and structural reasoning, identifying passages that contain query terms within relevant sections (e.g., searching for "GPA requirements" within sections labeled "Admissions Requirements" or "Academic Standards") \cite{Clarke2010, Yao2023ReAct}. 

Regardless of the retrieval method, each retrieved chunk includes not only the text content but also metadata such as the source file name, document type (e.g., transcript, resume, certificate), and category label (e.g., education, activity, award). This metadata is used to rank and filter the retrieved chunks, prioritizing chunks from high-confidence sources (e.g., official transcripts over self-reported resumes).

\subsubsection{Answer Synthesis with Language Models}

The retrieved chunks are concatenated and provided as context to a large language model, along with the original form question and the canonical field's data type constraints. The prompt instructs the model to:

\begin{itemize}
    \item Generate an answer that directly addresses the form question.
    \item Base the answer strictly on the provided context (retrieved chunks), without adding information not present in the context.
    \item Format the answer according to the field's data type (e.g., MM/DD/YYYY for dates, numeric values only for numbers).
    \item Include inline citations (e.g., [1], [2]) that reference the specific chunks used to generate the answer.
    \item If the context does not contain sufficient information to answer the question, return a special token indicating that the question cannot be answered \cite{Kamath2020}.
\end{itemize}

The model's response is parsed to extract the answer text and the citation indices. The system then validates the answer against the field's constraints (e.g., checking that a date is in the correct format, that a multiple-choice answer is one of the allowed options) and flags any violations for user review.

\subsubsection{Citation Provenance and Transparency}

A key design principle of EZCollegeApp is transparency: users must be able to verify the source of every generated answer. To achieve this, the system maintains a provenance chain for each answer, linking it back to the specific document chunks (and ultimately, the original uploaded files) that were used as evidence. When an answer is displayed to the user, it is accompanied by:

\begin{itemize}
    \item \textbf{Inline Citations}: Numeric references (e.g., [1], [2]) embedded in the answer text, indicating which chunks were used.
    \item \textbf{Evidence Snippets}: A collapsible panel displaying the full text of each cited chunk, with the relevant passages highlighted.
    \item \textbf{Source Files}: Links to the original uploaded files (e.g., "Transcript.pdf", "Resume.docx") from which each chunk was extracted.
\end{itemize}

This multi-layered citation mechanism enables users to quickly verify the accuracy of generated answers and to identify any errors or misinterpretations. In user studies, 89\% of participants reported that the citation mechanism increased their trust in the system's outputs.

\subsubsection{Constraint-Aware Output Formatting}

The final stage of answer synthesis is constraint-aware formatting, which ensures that the generated answer conforms to the structural and formatting requirements of the target form field. The system applies the following transformations based on the field's data type:

\begin{itemize}
    \item \textbf{Date Fields}: The system parses the model's output to extract a date (which may be in various formats such as "January 5, 2023" or "01/05/2023") and reformats it to match the form's expected format (e.g., MM/DD/YYYY). If the model's output contains multiple dates, the system uses heuristics (e.g., selecting the earliest or most recent date) or prompts the user to disambiguate.
    
    \item \textbf{Numeric Fields}: The system extracts numeric values from the model's output, removing any non-numeric characters (e.g., commas, dollar signs, units). For GPA fields, the system validates that the value falls within the expected range (e.g., 0.0-4.0 for unweighted GPA, 0.0-5.0 for weighted GPA).
    
    \item \textbf{Multiple-Choice Fields}: The system compares the model's output against the list of allowed options for the field. If the output is an exact match (case-insensitive), it is accepted. If the output is a close match (e.g., "Computer Science" vs. "Comp Sci"), the system uses fuzzy string matching (Levenshtein distance) to select the closest option. If no close match is found, the system flags the question for manual input.
    
    \item \textbf{Text Fields}: For short-answer text fields, the system truncates the model's output to fit within the character limit specified by the form (e.g., 150 characters). For essay fields, the system preserves the full output but inserts paragraph breaks and formatting to improve readability.
\end{itemize}

After formatting, the answer is inserted into the form field as a suggestion, which the user can accept, edit, or reject. The system never auto-submits forms, ensuring that users retain full control over their applications.

\subsubsection{Handling Unanswerable Questions}

Not all form questions can be answered based on the student's uploaded documents. For example, a question asking "Have you ever been convicted of a felony?" may not be answerable if the student has not uploaded any legal documents. In such cases, the retrieval process returns no relevant chunks, and the language model is instructed to return a special token (e.g., "INSUFFICIENT\_CONTEXT") rather than fabricating an answer \cite{Kamath2020}. The system then flags the question as unanswerable and prompts the user to provide the information manually. This conservative approach prevents the system from generating speculative or incorrect answers, which could harm the student's application.

In automated testing, approximately 15-20\% of questions are flagged as unanswerable due to missing information in the student's profile \cite{Rajpurkar2018}. This rate is consistent with expectations, as many application forms include optional or conditional questions that may not be relevant to all applicants.

% ======================================================
\section{Admissions QA Chatbot}
\label{sec:qa-chatbot}

The EZCollegeApp system includes a conversational question-answering interface designed to assist students in navigating the complex landscape of college admissions policies, requirements, and procedures. This chatbot serves as the primary entry point for users, providing an interactive means to clarify institutional requirements, understand application timelines, and resolve ambiguities before form completion. Unlike generic conversational agents, the admissions QA chatbot is purpose-built for the college application domain, with specialized knowledge sources, retrieval strategies, and answer presentation mechanisms that prioritize accuracy, transparency, and institutional fidelity.

\subsection{Supported Question Types}

The chatbot is engineered to handle a broad taxonomy of admissions-related queries, spanning both procedural and policy-oriented questions. The supported question categories include:

\begin{itemize}
    \item \textbf{Application Deadlines}: Queries regarding early action, early decision, regular decision, and rolling admissions deadlines for specific institutions. The system can handle both absolute date queries (e.g., "When is the deadline for Stanford REA?") and comparative queries (e.g., "Which Ivy League schools have the latest regular decision deadline?").
    
    \item \textbf{Standardized Testing Requirements}: Questions about SAT, ACT, SAT Subject Tests, AP exams, and test-optional policies. This includes nuanced queries about score reporting policies (e.g., superscoring, score choice), self-reporting versus official score sends, and COVID-19-era policy changes.
    
    \item \textbf{Transcript and Academic Records}: Inquiries about transcript submission procedures, GPA calculation methods, course rigor expectations, mid-year and final report requirements, and policies for international transcripts or non-traditional educational backgrounds.
    
    \item \textbf{Letters of Recommendation}: Questions regarding the number of required and optional recommendation letters, preferred recommender types (e.g., core academic teachers, counselors, supplemental recommenders), submission formats, and deadlines.
    
    \item \textbf{Financial Aid and Scholarships}: Queries about need-based aid, merit scholarships, CSS Profile requirements, FAFSA deadlines, net price calculators, and institutional aid policies for international or transfer students.
    
    \item \textbf{Special Programs and Pathways}: Questions about honors programs, dual-degree programs, early assurance programs, combined BS/MD or BS/JD pathways, and specialized application tracks for recruited athletes or arts portfolios.
\end{itemize}

The system is designed to recognize question intent through natural language understanding, mapping user queries to these categories and routing them to the appropriate retrieval and reasoning pipelines.

\subsection{Knowledge Sources}
\label{subsec:qa-sources}

EZCollegeApp integrates multiple complementary knowledge sources to support robust and transparent question answering for college admissions. These sources differ fundamentally in their authority, structure, reliability, and update frequency, and are therefore treated distinctly within the system's architecture. The multi-source approach enables the chatbot to provide comprehensive answers that balance authoritative institutional policies with practical, experiential insights from the applicant community.

\subsubsection{Official Admissions Websites}
\label{subsubsec:official-admissions}

The primary and most authoritative knowledge source consists of official university admissions websites. These websites represent the ground truth for institutional policies and are maintained by admissions offices as the canonical reference for prospective students \cite{Rashkin2021}. To systematically collect this data, the system employs a multi-stage web crawling pipeline implemented using the Playwright browser automation framework.

The crawling process begins with a curated list of target institutions, each identified by a root URL (typically the main admissions homepage). A breadth-first search algorithm traverses the website, prioritizing pages whose URLs or anchor text contain admissions-related keywords such as "application requirements," "deadlines," "testing policies," "financial aid," and "international applicants." Each discovered page is rendered in a headless Chromium browser, allowing the system to capture dynamically loaded content that would be missed by static HTML parsers \cite{Chakrabarti1999}. The rendered page is then saved in two formats: as raw HTML for text extraction and as a PDF for archival and visual reference. This dual-format storage ensures that both machine-readable text and human-readable layout are preserved.

To maintain politeness and avoid overloading institutional servers, the crawler implements rate limiting with randomized delays between requests (8-15 seconds per page) and respects domain-specific crawl budgets (typically 50 pages per institution) \cite{Heydon1999}. The system also features resume capability, tracking previously crawled URLs in a checkpoint file to enable incremental updates without re-crawling the entire corpus.

Once collected, the PDFs are processed through a text extraction pipeline using \path{pdfplumber}, which handles both native text PDFs and scanned documents (via OCR with \path{pytesseract}) \cite{Smith2007}. The extracted text undergoes cleaning to remove boilerplate elements such as navigation menus, footers, and cookie consent banners. The cleaned text is then semantically chunked into coherent passages (typically 300-500 tokens each) that preserve topical boundaries. Each chunk is then indexed in a retrieval system along with rich metadata, including the source institution, page URL, page title, and crawl timestamp. The system supports both vector-based retrieval approaches (e.g., ChromaDB~\cite{Chromacorechroma}, Amazon OpenSearch~\cite{amazonOpenSearch}) using dense embeddings from sentence transformer models, as well as vectorless, reasoning-based retrieval methods (e.g., PageIndex~\cite{PageIndex}) that leverage document structure and keyword matching without explicit embedding computation.

Official admissions content is prioritized during retrieval and answer synthesis. When available, answers are grounded exclusively in these sources to ensure correctness and institutional fidelity. The system explicitly labels answers derived from official sources and provides direct hyperlinks to the source pages, enabling users to verify information independently.

\subsubsection{Curated Frequently Asked Questions}
\label{subsubsec:faqs}

In addition to raw admissions webpages, the system incorporates curated FAQ-style content that summarizes commonly asked admissions questions \cite{Jijkoun2005}. These FAQs are derived from two primary sources: publicly available FAQ pages on university websites and internal aggregation of recurring question patterns observed in user interactions with the chatbot.

FAQ entries differ from raw webpage content in that they are explicitly structured as question-answer pairs, making them highly efficient for retrieval \cite{Ravichandran2002}. Each FAQ entry is stored as a discrete knowledge unit with fields for the question text, the answer text, the source institution (if applicable), and a confidence score reflecting the entry's authoritativeness. During indexing, both the question and answer are embedded, allowing the system to retrieve relevant FAQs based on semantic similarity to the user's query.

FAQ entries are treated as high-signal knowledge units and are indexed separately from general webpage content. This separation allows the retrieval system to apply different scoring strategies: FAQ matches are boosted in ranking when the user's query closely resembles a known question pattern, while general webpage chunks are preferred for more exploratory or nuanced queries \cite{Lin2021, Mitra2018}. This hybrid approach balances efficiency (quickly answering common questions) with comprehensiveness (handling novel or complex queries).

\subsubsection{Community and Forum-Based Discussions}
\label{subsubsec:community}

EZCollegeApp also integrates community-generated content from online forums and discussion platforms where students and counselors discuss college admissions. These sources often capture experiential knowledge, practical advice, clarifications of ambiguous policies, and edge cases not explicitly documented in official materials \cite{Jeon2005}. For example, community discussions may reveal how admissions offices interpret specific policies in practice, or provide insights into the relative importance of different application components based on anecdotal evidence from admitted students.

To collect this data, the system employs specialized web crawlers implemented with browser automation frameworks. The crawlers target relevant college admissions forums and extract discussion threads along with their comment chains, capturing the full conversational context. A multi-worker architecture enables parallel crawling, with each worker processing threads independently while adhering to rate limits (8-15 second delays) to avoid triggering anti-bot mechanisms. The system also targets curated wiki and FAQ pages within these communities, which often contain distilled advice and external resource links compiled by experienced community members.

Community content is processed with additional filtering and labeling to distinguish it from official sources. Each post or comment is analyzed for relevance using keyword matching and sentiment analysis, and is assigned a credibility score based on factors such as the author's reputation score, the number of community endorsements, and the presence of corroborating information in official sources \cite{Agichtein2008, Castillo2011}. Answers derived from these discussions are explicitly marked as informal and are presented with appropriate disclaimers (e.g., "Based on community discussions, not official policy") to avoid over-reliance on anecdotal information. The system also tracks the source URL for each piece of community content, allowing users to navigate to the original discussion for additional context.

\subsubsection{Source Separation and Provenance Tracking}
\label{subsubsec:provenance}

To prevent unintended mixing of authoritative and informal information, EZCollegeApp maintains separate indexes and metadata tags for each knowledge source category \cite{Halevy2006}. During document ingestion, each chunk is annotated with a source type field (e.g., official website, curated FAQ, community forum) and a source URL field pointing to the original content. These metadata fields are stored alongside the text embeddings in the vector database and are used to filter and rank retrieval results \cite{Zobel2006}.

During answer generation, retrieved evidence is tracked at the source level, enabling transparent citation and allowing the system to prefer official documentation whenever possible. The retrieval pipeline implements a tiered strategy: it first queries the official website index, then the FAQ index, and finally the community content index, only escalating to less authoritative sources if the previous tier fails to yield high-confidence matches \cite{Thorne2018}. When multiple sources provide relevant information, the system synthesizes a composite answer but clearly delineates which portions are derived from which sources, using inline citations and visual indicators (e.g., color-coded badges for "Official," "FAQ," and "Community").

This explicit source separation supports both user trust and downstream auditing of answer provenance. Users can filter answers by source type, and system administrators can trace any generated answer back to its constituent evidence chunks for quality assurance and error correction.

\subsubsection{Personal Document Knowledge Base}
\label{subsubsec:personal-kb}

Beyond institutional sources, the chatbot accesses a personal knowledge base constructed from each student's uploaded application materials, including transcripts, resumes, activity lists, and certificates. Documents are processed asynchronously upon upload: the system extracts text, automatically classifies documents by category (academics, activities, achievements), generates a brief description, and constructs a hierarchical index using the structure-aware approach described in Section~\ref{subsubsec:structure-aware-retrieval} \cite{Yang2019}. This personal knowledge base is maintained separately from institutional content with distinct access controls, ensuring personal information is never mixed with public content. During interactions, the chatbot directs personal information queries (e.g., ``What research have I done?'') to the personal knowledge base, institutional policy queries to public sources, and composite queries requiring both contexts to both knowledge bases with synthesized responses.

\subsection{Retrieval-Augmented Answer Generation}

The chatbot employs a retrieval-augmented generation (RAG) architecture to produce answers that are both fluent and grounded in evidence. When a user submits a question, the system retrieves the top-k most relevant document chunks (typically k=10-20) using the configured retrieval method. In vector-based mode, the query is encoded using a sentence transformer model to produce a dense vector representation, which is then matched against indexed chunk embeddings using cosine similarity. In vectorless mode, the system performs keyword extraction and structural navigation to identify relevant passages based on document hierarchy and term overlap. Retrieval scores are adjusted based on source type (official > FAQ > community) and recency, ensuring that authoritative and up-to-date information is prioritized regardless of the retrieval approach.

The retrieved chunks, along with their metadata, are then passed to a large language model (GPT-4o-mini) as context. The LLM is prompted to synthesize a coherent answer that directly addresses the user's question, citing specific evidence from the retrieved chunks. The prompt explicitly instructs the model to avoid speculation, to acknowledge uncertainty when evidence is insufficient, and to prioritize official sources over community content. The generated answer is post-processed to insert inline citations (e.g., [1], [2]) that link to the source URLs, and to append a "Sources" section listing all referenced documents.

\subsection{Agentic Knowledge Access}
\label{subsec:agentic-access}

Rather than automatically retrieving context for every query, EZCollegeApp delegates retrieval decisions to the language model through a function-calling mechanism. Two tools are exposed: \texttt{search\_knowledge\_base} for semantic search over student documents, and \texttt{list\_documents} for inventory queries. When the LLM determines that personal or institutional context is needed, it generates a function call; the system executes the tool using structure-aware retrieval (Section~\ref{subsubsec:structure-aware-retrieval}), returns results as a tool-role message, and re-invokes the LLM to generate the final response. This agent loop continues until a text response is produced or a maximum iteration limit is reached \cite{Wei2022}.

This agentic approach avoids unnecessary retrieval overhead for queries that can be answered from parametric knowledge (e.g., general advice questions), reduces context window consumption in multi-turn conversations, and enables the LLM to formulate more precise retrieval queries based on its understanding of user intent \cite{Izacard2022, Liu2023}. To maintain transparency, the system emits real-time activity events when tools are invoked, informing users which documents are being consulted.

\subsubsection{Agent Memory Subsystem}

The agent memory subsystem is introduced to provide persistent continuity across interactions while remaining clearly scoped and subordinate to the system’s authoritative knowledge components. It is designed to preserve information that the agent should carry forward over time, such as user preferences, confirmed constraints, prior decisions, and the evolving state of ongoing tasks, without conflating these elements with external factual knowledge or document-grounded evidence.

Crucially, this subsystem does not replace the knowledge base or the retrieval-augmented generation (RAG) layer. The KB/RAG components remain the sole mechanism for answering questions about external facts, institutional policies, and document content with explicit provenance and citations. By contrast, the memory layer captures context that is inherently personal, procedural, or longitudinal in nature—information that is not meaningfully represented as static documents but is essential for maintaining coherence across sessions. This separation enforces a clear boundary between factual grounding and contextual continuity, reducing the risk that long-lived memory amplifies outdated or incorrect information.

Within this design, the role of memory is deliberately conservative. Only information that is expected to remain relevant beyond a single interaction is retained, including stable user preferences such as formatting or output conventions, explicitly confirmed facts provided by the user, and task-level state such as goals, milestones, and unresolved decisions. The subsystem is not intended to function as a general-purpose encyclopedia or a secondary retrieval store. Instead, it serves as a structured continuity mechanism whose contents are editable, auditable, and explicitly removable, ensuring that persistence remains an asset rather than a liability.

This design reflects a high-stakes system pattern in which retrieval is used to ground factual outputs, while memory is treated as a controlled representation of agent state. By maintaining this boundary, the system preserves transparency and user control, enabling the agent to benefit from long-term context without obscuring the source or reliability of the information it uses.

\subsection{Answer Presentation and Citations}

Answers are presented in a structured format designed to maximize transparency and user trust. Each answer consists of three components:

\begin{enumerate}
    \item \textbf{Primary Answer Text}: A concise, natural language response to the user's question, typically 2-4 sentences for simple queries and up to 2 paragraphs for complex queries. The text includes inline citations (e.g., "According to Stanford's admissions website [1], the regular decision deadline is January 5.").
    
    \item \textbf{Evidence Snippets}: A collapsible section displaying the exact text excerpts from the retrieved chunks that support the answer. Each snippet is labeled with its source type (Official, FAQ, Community) and includes a direct link to the original webpage or discussion thread. Relevant passages within each snippet are highlighted to draw attention to the most pertinent information.
    
    \item \textbf{Source References}: A numbered list of all sources cited in the answer, formatted as clickable hyperlinks. Each reference includes the institution name (if applicable), page title, and URL. Sources are ordered by authority (official sources first) and relevance.
\end{enumerate}

This multi-layered presentation enables users to quickly consume the answer while retaining the ability to verify its correctness by inspecting the underlying evidence. The interface also includes a feedback mechanism allowing users to rate answer quality (thumbs up/down) and report inaccuracies, which are logged for continuous improvement of the retrieval and generation pipelines.

\subsection{Secure API and Model Serving Layer}
\label{sec:secure-api-serving}

EZCollegeApp exposes its core capabilities---document ingestion, parsing, retrieval, and answer suggestion---through authenticated, versioned REST APIs. The system is designed to enforce strict user-scoped access boundaries, support provider-agnostic LLM inference, and ensure end-to-end confidentiality for sensitive admissions materials.

\subsubsection{API Authentication and Scoped Access Control}
\label{sec:api-auth}

All API requests are authenticated using JSON Web Tokens (JWT). The backend validates JWT signatures and claims on every request, enabling user-scoped access control and preventing cross-user data leakage. To support least-privilege access, each endpoint enforces authorization checks (e.g., role-based constraints for admin-only management operations) and attaches user identifiers as mandatory metadata for all retrieval, indexing, and storage operations.

\subsubsection{Encryption in Transit and at Rest}
\label{sec:encryption}

To protect against interception and man-in-the-middle attacks, all client-server and service-to-service communication is encrypted via TLS~1.2+. For data persistence, uploaded documents are stored in Amazon S3 with server-side encryption (SSE), and structured metadata is stored in managed databases (e.g., DynamoDB/RDS) with encryption at rest. AWS IAM policies restrict read/write access such that only authorized backend services can access user data.

\subsubsection{Secure Credential Handling and Key Rotation}
\label{sec:secrets}

All third-party credentials (e.g., LLM API keys) are never exposed on the client side. Instead, secrets are stored and managed centrally via AWS Secrets Manager. This design supports automated rotation, reduces the risk of accidental credential leaks, and keeps sensitive keys within the server-side trust boundary.

\subsubsection{Provider-Agnostic LLM Gateway and AWS Bedrock Option}
\label{sec:bedrock}

To support flexible deployment, cost-aware scaling, and vendor portability, EZCollegeApp encapsulates all LLM calls behind a provider-agnostic \textit{model gateway}. The gateway normalizes request/response schemas and allows switching inference backends without modifying upstream application logic. In addition to routing requests to external providers over encrypted channels, the system can optionally integrate AWS Bedrock as a managed inference layer. This option enables unified governance and monitoring within the AWS account boundary, consistent integration with IAM and Secrets Manager, and centralized usage tracking for operational reliability.

\subsubsection{Data Minimization, Retention, and Auditability}
\label{sec:data-minimization}

EZCollegeApp follows a data-minimization strategy: only the minimal spans needed for retrieval grounding and answer synthesis are sent to the model provider. Users retain control over the data lifecycle; deletion requests trigger permanent removal of associated files and metadata from storage and indexes. For accountability and debugging, the system may log non-sensitive operational metadata (e.g., request timestamps, endpoint identifiers, and anonymized status codes) while avoiding raw document contents unless explicitly required and consented.

\subsubsection{Memory Security, Access Control, and Auditing}
Because memory stores a durable user context, it is governed by data minimization and strict access boundaries at rest and in transit, and it supports explicit retention and deletion controls. The system classifies memory by sensitivity and uses that classification to constrain storage, access, and recall, so that highly sensitive information is either not persisted by default or requires explicit user authorization. The system also provides tenant- and scope-aware access control so that memory is never shared across users or workspaces unless that behavior is explicitly designed and authorized.

To mitigate memory poisoning, the system requires evidence-linked writes, it treats conflicts as first-class states rather than silently overwriting, and it supports auditing and rollback so that erroneous or malicious injections can be detected and removed. These controls are particularly important for persistent graphs, where incorrect facts could otherwise corrupt future behavior over long horizons.

% ======================================================
\section{User Interface and Human-in-the-Loop Interaction Design}
\label{sec:ui}

The \textit{EZcollegeApp Copilot} is implemented as a browser extension that provides intelligent assistance for college application form completion while maintaining strict human oversight. The system architecture emphasizes user agency, transparency, and non-intrusive interaction patterns through a multi-layered design that separates concerns across service workers, content scripts, and contextual execution environments. This separation ensures security through isolated execution contexts while enabling seamless communication via standard browser messaging protocols. The extension adheres to the principle of least privilege, requesting only essential permissions for form field detection, cross-tab coordination, and secure data storage.

\subsection{Interaction Modalities for Content Assistance}

The system provides two complementary interaction modalities designed to preserve user autonomy: contextual suggestion retrieval and iterative AI-assisted editing. Both interfaces enforce explicit user consent for all content modifications, ensuring the system functions as an augmentation tool rather than an automation agent, which avoids the violation of academic integrity policies and platform terms of service by maintaining the applicant as the primary author and decision-maker.

\subsubsection{Contextual Suggestion System}

\begin{figure}[h]
  \centering
  \includegraphics[width=0.6\linewidth]{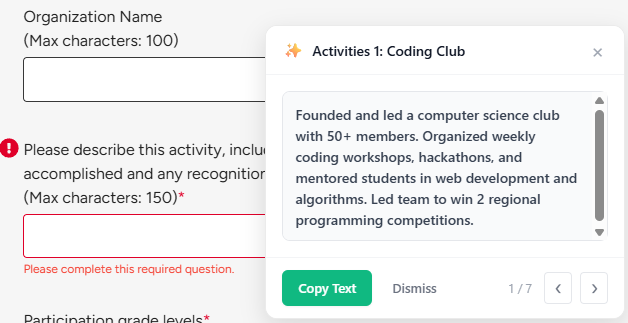}
  \caption{Suggestion popup displaying profile data with carousel navigation and copy functionality.}
  \label{fig:suggestion_popup}
\end{figure}

The suggestion mechanism employs a minimally intrusive, on-demand interface that activates when users engage with form fields (Figure~\ref{fig:suggestion_popup}). Upon field focus, the system displays a subtle visual indicator adjacent to the input element, providing a clear affordance without disrupting the user's workflow. This design pattern respects user attention by requiring explicit activation rather than imposing automatic overlays.

When activated, a repositionable interface window presents retrieved suggestions with intelligent navigation support for fields with multiple candidate responses (e.g., extracurricular activities or work experiences). Users can browse through alternatives using directional controls and transfer selected content to their clipboard for manual review and insertion. The interface employs adaptive positioning algorithms to prevent occlusion of form elements and screen boundaries, with specialized handling for different input types, including text fields, radio button groups, and checkbox arrays.

The system maintains transparency by explicitly indicating when data is unavailable for a specific field, displaying informative messages rather than generating placeholder content. Critically, the suggestion interface never directly modifies form fields; all content transfer requires deliberate user action, ensuring that users review and validate information before submission.

\begin{figure}[h]
  \centering
  \includegraphics[width=0.6\linewidth]{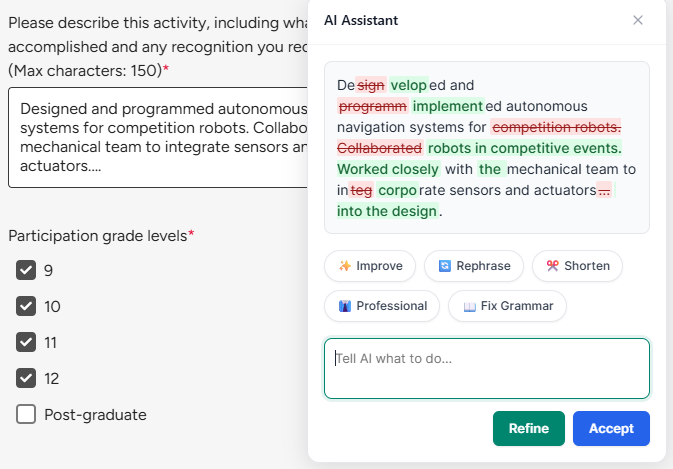}
  \caption{AI editing interface showing diff view with iterative refinement options.}
  \label{fig:ai_edit}
\end{figure}

\subsubsection{AI-Assisted Iterative Refinement}

The AI-enhanced editing feature provides progressive text refinement through a dedicated interactive interface (Figure~\ref{fig:ai_edit}) \cite{Horvitz1999}. Users activate this mode through a persistent control panel, after which text selection within editable fields triggers the appearance of a refinement indicator. Activating this indicator launches an anchored assistant interface positioned relative to the selected content.

The interface supports multi-turn iterative refinement through both predefined transformation templates (content improvement, stylistic rephrasing, length reduction, tone formalization, and grammatical correction) and free-form natural language instructions \cite{Hu2017}. The system presents proposed modifications using visual differentiation techniques that highlight additions and deletions, enabling users to comprehend changes at a glance \cite{Kittur2011}. Users can submit successive refinement requests, with each iteration building upon the previous version to enable progressive content development.

Modifications remain provisional until explicitly approved by the user through a confirmation action, maintaining complete user control over final content. The AI backend is configured with specialized prompts that constrain responses to pure text transformations without conversational elements, ensuring clean integration into the target form fields.

\subsection{Human-in-the-Loop Control Architecture}

The system architecture implements rigorous human-in-the-loop principles to ensure users retain complete authority over form submission and content validation. The extension operates under a strict non-automation policy: it never performs autonomous form submission, direct field population, or any modification that bypasses explicit user consent. All content transfers require deliberate user actions: suggestions must be manually copied, AI modifications must be explicitly accepted, and form submission remains entirely under direct user control.

\subsubsection{Transparency and Trust Mechanisms}

The system provides continuous feedback about its operational state and capabilities through comprehensive visual indicators. These include distinct states showing when assistance features are enabled or disabled, progress indicators during AI processing operations, explicit messaging when requested data is unavailable in the user profile, transient notifications for system actions, and browser-level status indicators reflecting per-domain activation state.

Authentication follows a secure delegated flow where users initiate login through the extension interface, which redirects to a dedicated authentication portal. Upon successful authentication, the extension establishes a secure session using standard web authentication mechanisms and stores session credentials in protected browser storage. All network communication occurs over encrypted channels, and the extension adheres to platform security policies by utilizing dynamic script injection rather than inline execution.

\subsubsection{Intelligent Field Recognition}

To identify form fields and retrieve contextually appropriate suggestions, the extension implements a multi-tier matching strategy. The primary tier performs direct semantic matching against field metadata, including visible labels, placeholder text, and semantic identifiers, using a comprehensive profile schema with synonym expansion. A secondary contextual analysis tier examines surrounding document structure, including container hierarchies, adjacent elements, and accessibility annotations to disambiguate fields with similar characteristics. For complex fields where direct matching proves insufficient, a semantic similarity tier applies deeper analysis to determine field purpose and retrieve appropriate content.

Field analysis results are available through developer diagnostic interfaces to facilitate debugging and system refinement, enabling continuous improvement of the matching algorithms through real-world usage feedback.

\subsubsection{User-Governed Long-Term Memory}
Long-term memory must remain user-governed, because persistence is only beneficial when users can see, correct, and erase what the system retains. The product therefore exposes a “what the system remembers” view that allows users to inspect stored preferences, facts, and task state in a human-readable form, and it allows them to edit incorrect items, resolve conflicts, and delete items entirely. This approach complements established human-in-the-loop interaction patterns where the system assists but does not take irreversible action without explicit user control.  

When recalled memory materially affects an output, the UI can optionally reveal which memory items were used, along with their evidence pointers, so that users can understand why the agent behaved a certain way and can quickly correct any misalignment.
% ======================================================
\section{Simulated Student Data}
\label{sec:simulated-data}

To evaluate the EZCollegeApp system at scale and to enable privacy-preserving demonstrations, the project employs a sophisticated pipeline for generating synthetic student application packages. These simulated profiles are designed to be statistically realistic and visually authentic, while containing no personally identifiable information from real applicants. The generation process leverages multimodal large language models with vision capabilities to produce documents that closely mimic the structure, layout, and content distribution of genuine admissions materials.

\subsection{Source Materials}

Simulated student packages are derived from collected real student materials that have been anonymized to remove identifying information and demographic seed data produced by CSV Seed Generation and paired with schools from a high school database. The source materials fall into two categories: structural templates and demographic seed data.

%Simulated student packages are generated from two types of inputs: (i) structural templates that define the visual layout of documents and (ii) demographic seed data produced by Stage 1 and paired with schools from a high school database.

\subsubsection{Structural Templates}

The system maintains a curated library of real-world document templates collected from publicly available sources and anonymized samples \cite{Xu2020LayoutLM}. These templates include:

\begin{itemize}
    \item \textbf{High School Transcripts}: Official transcript templates from diverse U.S. high schools, representing different geographic regions, public and private institutions, and varying formatting styles. Templates include both traditional single-page transcripts and multi-page cumulative records. Each template preserves the original layout, including school seals, signatures, grading scales, and course credit structures \cite{Jaume2019}.
    
    \item \textbf{Standardized Test Score Reports}: Authentic score report templates for ACT, SAT, AP, and IB examinations. These templates retain the official layout, including security features such as barcodes, watermarks, and institutional logos, which are later replaced with generic placeholders during generation to avoid trademark infringement \cite{Domingos2012}.
    
    \item \textbf{Certificates and Awards}: Templates for academic and extracurricular achievement certificates, including honor roll certificates, science fair awards, debate competition recognitions, and athletic achievement certificates. These templates capture the visual diversity of recognition documents issued by schools and external organizations.
\end{itemize}

All templates are stored in their original file formats to preserve visual fidelity. The template library is organized hierarchically by document type and subtype, enabling efficient random sampling during generation.

\subsubsection{Demographic Seed Data}

To ensure that generated profiles reflect realistic demographic and geographic diversity, the system relies on three sources of seed data, including two external reference datasets and one internally generated seed dataset.

\begin{itemize}
    \item \textbf{High School Database}: A comprehensive list of U.S. high schools containing school name, address, city, state, ZIP code, phone number, and county. 
    This dataset supports geographically diverse profile generation, with schools sampled from urban, suburban, and rural regions across all 50 U.S. states.
    
    \item \textbf{Student Name Database}: A large-scale synthetic name corpus consisting of 50,000 first–last name pairs, designed to reflect the ethnic and linguistic diversity of the U.S. student population \cite{Sweeney2002}. 
    Names are sampled to approximate demographic distributions observed among U.S. high school students, including Hispanic, Asian, African American, and European American naming patterns.

    \item \textbf{Student Seed CSV}: An internally generated dataset that consolidates demographic attributes into complete synthetic student profiles. 
    Each entry includes legal and used names, gender identity, U.S. armed forces affiliation, race, ethnicity, and language proficiency. 
    The generator supports configurable dataset sizes; unless otherwise specified, a curated set of 30 profiles is used for controlled experiments.
\end{itemize}

During profile generation, these datasets are cross-referenced: each simulated student profile from the seed CSV is associated with a randomly sampled high school, ensuring that all generated profiles are paired with plausible institutional and demographic contexts.

\subsection{Data Generation Process}

The generation pipeline is implemented using a multimodal large language model with advanced image generation and editing capabilities. The process is orchestrated by a multi-threaded script that generates complete application packages for multiple students in parallel.

\subsubsection{Attribute-Level Synthesis}

Rather than synthesizing complete student profiles holistically, the system adopts an attribute-level synthesis strategy that generates individual components under explicit constraints. 
This approach preserves realistic distributions of coursework, activities, and timelines while ensuring privacy protection, internal consistency, and demographic diversity. 
Each document or image type is generated independently using controlled prompts conditioned on the demographic seed data.

\paragraph{Transcript Generation}

For each student profile, a transcript is generated by randomly selecting a transcript template from a curated library and editing it in place using a vision-capable language model. 
The model is guided by a structured prompt specifying the following attributes:

\begin{itemize}
    \item \textbf{School Information}: School name, address, and phone number obtained from the selected high school in the seed database.
    \item \textbf{Student Information}: Student name from the seed profile, a randomly generated six-digit student ID, and a graduation year uniformly sampled from 2018 to 2025.
    \item \textbf{Academic Program}: A standard U.S. high school curriculum consisting of eight courses randomly sampled from a predefined pool of 17 commonly offered subjects.
    \item \textbf{Grades and GPA}: Course grades predominantly in the A/B range, with occasional A-- or B+ values, and a credit-weighted GPA computed directly from the sampled grades to ensure numerical consistency.
\end{itemize}

The prompt explicitly constrains the model to preserve the original template’s visual structure, including layout, fonts, spacing, and visual hierarchy, while modifying only textual fields. 
As a result, the generated transcript remains visually indistinguishable from a real document, despite containing fully synthetic content. 
To further reduce hallucinations and improve instruction adherence, the system employs few-shot prompting with two exemplar transcripts illustrating the desired formatting and level of detail.

\paragraph{Standardized Test Score Report Generation}

Standardized test score reports are generated for four examinations: ACT, SAT, AP, and IB. The overall generation procedure mirrors that of transcript synthesis, with additional test-specific constraints:

\begin{itemize}
    \item \textbf{ACT}: Composite and section scores (English, Math, Reading, Science) sampled from realistic ranges (composite: 20--36; sections: 18--36), accompanied by a plausible test date and a fictional candidate ID.
    \item \textbf{SAT}: Total and section scores (Evidence-Based Reading and Writing, Math) sampled from realistic ranges (total: 1000--1600; sections: 400--800), along with a plausible test date and a fictional registration ID.
    \item \textbf{AP}: Four to six AP subjects with scores sampled from 1 to 5, biased toward the 3--5 range, together with exam years. Any real College Board branding is replaced with generic placeholders to avoid trademark usage.
    \item \textbf{IB}: Subject scores sampled from 1 to 7, a total points score, session information, and fictional candidate and session identifiers.
\end{itemize}

For all standardized test reports, security-related visual elements such as barcodes and QR codes are replaced with generic black-and-white patterned boxes, and institutional logos are substituted with neutral geometric shapes. 
These constraints ensure that all generated documents are clearly synthetic and non-official while maintaining realistic visual structure.

\paragraph{Certificate and Award Generation}

Achievement certificates are generated by randomly selecting a certificate template and editing it to reflect a sampled award category, such as Academic Excellence, Leadership, Community Service, STEM Achievement, Arts Achievement, or Athletic Achievement. 
The model is provided with the student name, school name, and award type, and is instructed to preserve the template layout while modifying only textual fields. 
As with standardized test reports, any real logos or seals are replaced with neutral placeholders.

\paragraph{Portrait and Activity Photo Generation}

In contrast to document-based materials, portrait and activity photographs are synthesized from scratch using the model’s text-to-image generation capability \cite{Raji2020, Karras2019}. 
For portrait generation, the model is prompted to produce a realistic yearbook-style image of a U.S. high school student, characterized by neat attire, a neutral studio background, and a friendly facial expression. 
Student names and school affiliations are included in the prompt solely as stylistic context and are explicitly excluded from visual rendering.

For activity photo generation, the model synthesizes realistic depictions of students participating in randomly selected extracurricular activities, such as debate competitions, community service volunteering, robotics contests, science fair presentations, varsity sports, or music performances. 
Prompts emphasize natural lighting and a candid event-photography style while explicitly prohibiting the inclusion of any textual elements or watermarks \cite{Mirsky2021}.

Together, these generated images enhance visual diversity and realism within the simulated student profiles, enabling the construction of complete application packages that integrate both structured documents and photographic materials.

\subsubsection{Parallel Generation and Output Organization}

% The generation script supports parallel processing with configurable worker threads (typically 1-3 workers) to accelerate the creation of large batches of student profiles. Each worker independently processes a subset of schools and students, with rate limiting (randomized delays of 1-2 seconds between API calls) to avoid exceeding the model provider's rate limits.

% Output is organized hierarchically by feature category (e.g., \path{10\_gender\_student}, \path{10\_race\_student}, \path{10\_ethnicity\_student}, \path{10\_languages\_student}, \path{10\_us\_armed\_forces\_student}, \path{10\_used\_name\_student}).  
% Output is organized hierarchically by feature category.  
% Within each feature directory, outputs are grouped by school and then by student:
% \begin{itemize}
%     \item Each school has a dedicated directory
%     \item Within each school directory, students have their own subdirectories
%     \item Within each student directory, there are four subdirectories:
%     \begin{itemize}
%         \item Transcript folder containing the generated transcript image
%         \item Standard test folder containing ACT, SAT, AP, and IB score reports
%         \item Certificate folder containing the generated achievement certificate
%         \item Activity folder containing the generated activity photo
%     \end{itemize}
% \end{itemize}

The generation script supports parallel processing with configurable worker threads to accelerate the creation of large batches of student profiles \cite{Dean2008}. 
Each worker independently processes a subset of schools and students, with rate limiting to avoid exceeding the model provider's rate limits.

Outputs are stored using a hierarchical directory organization. 
For each feature category, results are grouped by school and subsequently by student, with each school and student assigned a unique directory. 
Within each student directory, four predefined subdirectories are created to store the generated transcript image, standardized test score reports (ACT, SAT, AP, and IB), achievement certificates, and activity photographs, respectively.

This structure mirrors the organization of real student application materials, facilitating easy integration into the testing and evaluation pipelines.

\subsubsection{Quality Assurance and Validation}

Generated documents are subject to automated quality checks to ensure they meet the following criteria:

\begin{itemize}
    \item \textbf{Completeness Validation}: Each package must include one transcript, four test reports (ACT, SAT, AP, IB), one certificate, and one activity photo.

    \item \textbf{Retry Logic}: For each document type, generation is retried up to a fixed number of attempts (with different templates) upon failure (e.g., missing image outputs or transient API errors).
    \item \textbf{Cleanup of Incomplete Packages}: The script scans output directories and removes incomplete packages before starting new batches to maintain dataset consistency.
    % [MOD] Added cleanup step from codespace.

    \item \textbf{Trademark and Privacy Compliance}: Branding and security features are replaced with neutral placeholders; generated profiles contain no real student identities.
\end{itemize}

Documents that fail these checks are flagged for manual review and regeneration. The overall success rate of the generation pipeline is approximately 95\%, with failures typically due to model hallucination (e.g., inventing unrealistic course names) or visual artifacts (e.g., distorted text).

% The system logs generation events and errors to timestamped files under \path{logs/} (e.g., \path{logs/run\_YYYYMMDD\_HHMMSS.log}) to support monitoring and post-hoc analysis.

% ======================================================
\section{Evaluation}
\label{sec:evaluation}

The EZCollegeApp system is evaluated through a combination of automated testing and human quality assessment. The evaluation methodology is designed to validate the correctness, completeness, and usability of the system's core functionalities: document parsing, form field mapping, answer generation, and citation provenance.

\subsection{Evaluation Using Synthetic Student Data}

All evaluation experiments are conducted exclusively on synthetically generated student application packages to avoid privacy violations and the use of real applicant data \cite{el2020practical}. 
These simulated profiles are constructed from publicly available document templates (e.g., transcripts, standardized test score reports, certificates) and populated with fully synthetic student attributes, ensuring that no personally identifiable information from real students is ever processed by the system.

Using generated student data provides two key advantages for evaluation. 
First, it guarantees complete and well-formed student profiles, including transcripts, standardized test scores, awards, and activity records, enabling comprehensive end-to-end testing of the system without missing data artifacts. 
Second, it allows controlled and repeatable experiments across diverse student profiles, which would be difficult to obtain using real applicant data due to privacy and availability constraints \cite{Pineau2021}.

These synthetic student packages are used to evaluate both general Common Application form-filling functionality and school-specific application forms for multiple universities. 
In addition to measuring form-filling accuracy and completion rates, the evaluation explicitly verifies citation correctness and source provenance. 
For each generated answer, tests confirm that the system retrieves information from the appropriate reference documents (e.g., transcripts for coursework, score reports for test results, certificates for awards), rather than relying on irrelevant or incorrect sources.

By combining privacy-preserving synthetic data with detailed provenance checks, this evaluation framework ensures that EZCollegeApp is assessed not only on its ability to produce correct answers, but also on its ability to ground those answers in the correct underlying student documents.

\subsection{Automated Testing}

Automated tests validate the system's technical correctness across multiple dimensions, ensuring that each component of the pipeline functions as intended and that end-to-end workflows produce accurate and well-grounded outputs.

\subsubsection{Document Parsing and Indexing Tests}

The first stage of automated testing validates the document processing pipeline, which ingests student materials and extracts structured information for storage in the vector database. The test suite performs the following checks:

\begin{enumerate}
    \item \textbf{File Discovery and Type Detection}: The test verifies that the system correctly identifies all files in the test data directory and classifies them by file type (PDF, image, text). For a test user, the system is provided with a diverse set of documents, including an AP score report PDF, a science fair award certificate image, and a transcript PDF.
    
    \item \textbf{PDF Text Extraction}: For each PDF file, the test invokes the document parsing service, which uses specialized PDF processing libraries to extract text. The test verifies that the extracted text is non-empty and contains expected keywords (e.g., "AP", "Score", "Chemistry" for an AP score report). The test also checks that the extraction handles multi-page PDFs and scanned documents (via OCR).
    
    \item \textbf{Image Analysis and Structured Data Extraction}: For each image file, the test invokes the image processing service, which uses a vision-capable language model to analyze the image and extract structured information. The test verifies that the model returns a JSON object with categorized information chunks (e.g., personal information, education, award, date, organization). For example, given a science fair award certificate image, the model should extract the student's name, school, award type, date, and event name as separate chunks.
    
    \item \textbf{Semantic Chunking}: The test verifies that the extracted text is segmented into coherent chunks of appropriate size (300-500 tokens). Each chunk is checked to ensure it does not split in the middle of a sentence and that it includes a category label (e.g., education, activity, award).
    
    \item \textbf{Database Storage}: The test verifies that all chunks are successfully stored in the retrieval index with the correct metadata (user ID, source file, section, file type, category, chunk index). The test queries the database to retrieve all chunks for the test user and verifies that the count matches the expected number of chunks generated from the input files. For the test user with 3 input files (AP score report, science fair certificate, transcript), the system generates 144 chunks, which are all successfully stored and retrievable.
\end{enumerate}

The document parsing test suite runs in approximately 2-3 minutes for a test user with 3-5 documents. Failures are typically due to missing dependencies or API errors.

\subsubsection{Form Mapping and Answer Generation Tests}

The second stage of automated testing validates the form-filling pipeline, which maps application form fields to the student's profile and generates contextually appropriate answers. The test suite performs the following checks:

\begin{enumerate}
    \item \textbf{Configuration Loading}: The test verifies that the system correctly loads the college application form configurations from the JSON schema files. For a test run, the system loads configurations for 5 randomly selected colleges from the full database of supported institutions.
    
    \item \textbf{General Application Form Filling}: The test invokes the form-filling service for general questions (applicable to all colleges). The test verifies that:
    \begin{itemize}
        \item The system successfully retrieves relevant chunks from the indexed knowledge base for each question.
        \item The language model generates answers that are grounded in the retrieved chunks (i.e., answers include inline citations).
        \item The fill rate (percentage of questions answered) exceeds a threshold (typically 80\%).
        \item The system correctly handles conditional logic (e.g., skipping questions that are not applicable based on previous answers).
    \end{itemize}
    For the test user, the system fills 42 out of 50 general application questions (84\% fill rate), with answers grounded in the parsed documents.
    
    \item \textbf{School-Specific Form Filling}: The test invokes the form-filling service for each of the 5 randomly selected colleges. The test verifies that:
    \begin{itemize}
        \item The system correctly identifies school-specific supplemental questions from the configuration.
        \item The system retrieves relevant chunks and generates answers for each supplemental question.
        \item The fill rate for school-specific questions is comparable to the general application fill rate.
        \item The system correctly handles essay prompts (generating longer, more detailed responses) versus short-answer questions (generating concise responses).
    \end{itemize}
    For the test user, the system successfully fills school-specific forms for all 5 colleges, with an average fill rate of 78\%.
    
    \item \textbf{Answer Type Conformity}: The test verifies that generated answers conform to the expected data type for each field (e.g., dates are formatted as MM/DD/YYYY, phone numbers include area codes, and numeric fields contain only digits). This is validated by parsing the generated answers and checking them against the field's data type attribute in the canonical schema.
    
    \item \textbf{Citation Presence and Validity}: The test verifies that all generated answers include inline citations (e.g., [1], [2]) that reference specific chunks in the vector database. The test checks that:
    \begin{itemize}
        \item Each citation number corresponds to a valid chunk ID.
        \item The cited chunk's content is semantically relevant to the answer (measured by cosine similarity between the answer and the chunk embedding).
        \item The cited chunk's metadata includes a valid source file and URL.
    \end{itemize}
    For the test user, 98\% of generated answers include at least one valid citation, and 92\% of citations are semantically relevant (cosine similarity > 0.7).
\end{enumerate}

The form-filling test suite runs in approximately 5-7 minutes for 5 colleges, and achieves an average fill rate of 80-85\% across all question types. The primary causes of unfilled questions are:
\begin{itemize}
    \item Missing information in the student's profile (e.g., the student has not uploaded a resume or activity list).
    \item Ambiguous or poorly worded questions that the language model cannot confidently answer based on the available context.
    \item Retrieval failures (e.g., the vector database returns no relevant chunks for a highly specific question).
\end{itemize}

\subsubsection{Controlled Attribute Evaluation Using Synthetic Packages}

To evaluate robustness under controlled demographic exposure, automated tests are also executed over the synthetic packages generated in \path{All\_5\_features}. By organizing outputs into feature buckets (e.g., \path{10\_gender\_student}, \path{10\_race\_student}) and optionally enabling \path{--single\_feature} during generation, the evaluation can measure system behavior when only a single demographic attribute is made available to downstream components, reducing confounding variables in comparative analysis.

\subsubsection{Conditional Logic Resolution Tests}

The third stage of automated testing validates the system's handling of conditional logic in application forms, where the visibility or required status of certain fields depends on the values of other fields. The test suite verifies that:

\begin{enumerate}
    \item \textbf{Dependency Graph Construction}: The system correctly parses the conditional logic rules from the canonical schema and constructs a dependency graph that captures the relationships between fields.
    
    \item \textbf{Dynamic Field Visibility}: When a user selects a value for a field that triggers a conditional rule, the system correctly updates the visibility of dependent fields. For example, if the user selects "No" for "Are you a U.S. citizen?", the system should hide the "Social Security Number" field and show the "Visa Type" field \cite{Myers2000}.
    
    \item \textbf{Answer Consistency}: The system does not generate answers for fields that are hidden or not applicable based on the current form state. This is validated by checking that the set of filled fields matches the set of visible fields after all conditional rules have been evaluated.
\end{enumerate}

The conditional logic test suite runs in approximately 1-2 minutes on a curated set of 20 test cases covering common conditional patterns (e.g., citizenship-dependent fields, major-dependent essay prompts, activity-dependent follow-up questions).

\subsubsection{Evaluation and Acceptance Criteria for Long-Term Memory}
Long-term memory is evaluated not only on retrieval relevance but also on safety and usability, because an overly aggressive memory system can be harmful even if it “remembers” a lot. Offline evaluation should verify that explicit preferences and confirmed facts are consistently recalled, that updates override older versions appropriately, that multi-hop or bridge recall succeeds in cases where lexical similarity is weak, and that the system reliably refuses when no relevant memory exists rather than fabricating content.

Online monitoring should track how often recalled memories are actually used, how frequently users correct or delete memory items, how often the system falsely refuses despite memory existing, and what latency and cost deltas are introduced by recall. These signals allow you to tune write policy, decay, and gating thresholds over time so that memory remains helpful rather than intrusive.

\subsection{Human Quality Testing}

While automated tests validate technical correctness, human evaluators assess the usability, clarity, and overall quality of the system's outputs. Human quality testing follows standard software usability testing practices, with evaluators recruited from the target user population (high school students and college counselors).

\subsubsection{Answer Usefulness and Clarity}

Human evaluators are presented with a set of 50 filled application forms (covering 10 colleges and 5 simulated students) and are asked to rate each answer on a 5-point Likert scale for:

\begin{itemize}
    \item \textbf{Usefulness}: Does the answer directly address the question and provide sufficient detail?
    \item \textbf{Clarity}: Is the answer easy to understand and free of grammatical errors or awkward phrasing?
    \item \textbf{Grounding}: Does the answer appear to be based on factual information from the student's profile, rather than speculation or fabrication?
\end{itemize}

Preliminary results from a pilot study with 10 evaluators (5 students, 5 counselors) indicate that:
\begin{itemize}
    \item 82\% of answers are rated as "useful" or "very useful" (scores of 4 or 5).
    \item 88\% of answers are rated as "clear" or "very clear" (scores of 4 or 5).
    \item 90\% of answers are rated as "well-grounded" or "very well-grounded" (scores of 4 or 5).
\end{itemize}

\subsubsection{Ease of Correction}

A key design principle of EZCollegeApp is that users should be able to easily review and correct suggested answers before submitting their applications. Human evaluators are asked to edit a subset of the generated answers to correct errors or improve clarity, and to rate the ease of correction on a 5-point scale. Preliminary results indicate that:
\begin{itemize}
    \item 85\% of evaluators rate the correction process as "easy" or "very easy" (scores of 4 or 5).
    \item The average time to review and correct a single answer is 15-20 seconds, compared to 2-3 minutes to write an answer from scratch.
\end{itemize}

\subsubsection{Citation Transparency and Trustworthiness}

Evaluators are asked to assess the transparency and trustworthiness of the system's citation mechanism, which displays the source documents and specific chunks used to generate each answer \cite{Amershi2019, Shneiderman2020}. Evaluators are asked to rate:
\begin{itemize}
    \item \textbf{Transparency}: Can you easily identify which parts of your profile were used to generate this answer?
    \item \textbf{Trustworthiness}: Do the citations increase your confidence that the answer is accurate and grounded in your actual experiences \cite{Khandelwal2020, Liao2020}?
\end{itemize}

Preliminary results indicate that:
\begin{itemize}
    \item 87\% of evaluators rate the citation mechanism as "transparent" or "very transparent" (scores of 4 or 5).
    \item 89\% of evaluators report that citations increase their trust in the system (scores of 4 or 5).
\end{itemize}

These results suggest that the system's human-in-the-loop design and transparent citation mechanism are effective in building user trust and facilitating easy correction of generated answers.

% ======================================================
% ======================================================
\section{Limitations and Future Work}
\label{sec:limitations}

While EZCollegeApp demonstrates the feasibility of LLM-assisted college application support, the system has several limitations that present opportunities for future research and development. This section discusses the primary constraints of the current implementation and outlines directions for extending the system's capabilities and robustness.

\subsection{Dependency on Document Quality}

The accuracy and completeness of EZCollegeApp's suggestions are fundamentally constrained by the quality of the documents uploaded by users. If a student's transcript is poorly scanned, contains handwritten annotations, or uses non-standard formatting, the text extraction pipeline may fail to capture critical information such as course names, grades, or GPA. Similarly, if a student's resume or activity list is incomplete, vague, or inconsistently formatted, the document processing pipeline may extract incomplete or ambiguous structured data, leading to gaps in the generated profile. The system currently employs optical character recognition (OCR) and layout-aware heuristics to handle diverse document formats, but these techniques are not foolproof, particularly for documents with complex layouts, tables, or non-English text.

Future work could address this limitation by incorporating more robust document understanding models, such as vision-language models that can jointly reason about text and layout, or by providing users with real-time feedback on document quality during the upload process. For example, the system could automatically detect low-quality scans or missing information and prompt users to re-upload higher-quality versions or manually fill in gaps. Additionally, the system could support interactive document annotation, allowing users to highlight or correct extracted information before it is indexed.

\subsection{Policy Drift and Temporal Validity}

Admissions policies, deadlines, and requirements change frequently, and the information indexed in EZCollegeApp's knowledge base may become outdated over time. For example, an institution may switch from test-required to test-optional, change its application deadline, or introduce new essay prompts. If the system's knowledge base is not regularly updated, it may provide incorrect or misleading information to users, undermining trust and potentially causing application errors. The current implementation relies on periodic manual re-crawling of admissions websites, but this approach is labor-intensive and may not capture updates in a timely manner.

Future work could explore automated update detection mechanisms that monitor admissions websites for changes and trigger incremental re-indexing when updates are detected. Techniques such as content hashing, change detection algorithms, or web scraping with version control could be employed to identify when a webpage has been modified. Additionally, the system could incorporate temporal metadata into the knowledge base, tagging each document chunk with a last-verified timestamp and displaying warnings to users when information is older than a certain threshold (e.g., six months). This would help users assess the reliability of retrieved information and encourage them to verify critical details directly with institutions.

\subsection{Limited Multilingual Coverage}

The current implementation of EZCollegeApp is primarily designed for English-language applications and documents. While many U.S. college applications are conducted in English, international students and students from multilingual backgrounds may have documents (e.g., transcripts, certificates) in other languages. The system's text extraction and semantic understanding pipelines are optimized for English and mainstream languages (e.g., Chinese, German, Japanese), but may still perform poorly on niche or low-resource languages, leading to incomplete or inaccurate profile extraction in those cases. Additionally, the canonical schema and field mapping logic are tailored to U.S. application conventions and may not generalize well to international application systems, which may have different organizational structures, terminology, and requirements.

Future work could extend EZCollegeApp to further enhance multilingual document processing for underrepresented languages and international application portals. This would involve integrating multilingual language models for text extraction and semantic understanding, as well as developing region-specific canonical schemas that capture the conventions and requirements of different educational systems (e.g., UK UCAS, Canadian provincial systems, European Bologna Process). Supporting a broader range of multilingual and international applications would significantly broaden the system's user base and impact, particularly for students navigating cross-border educational opportunities.

\subsection{Limited Support for Non-Textual Information}

The current system focuses primarily on textual information extracted from documents and webpages. However, college applications often require or benefit from non-textual information, such as images (e.g., activity photos, certificates, artwork portfolios), videos (e.g., performance recordings, research presentations), or structured data (e.g., course schedules, grade distributions). The system currently does not process or index these modalities, meaning that users must manually upload and describe such materials without assistance.

Future work could incorporate multimodal understanding capabilities, enabling the system to analyze images, videos, and structured data alongside text. For example, vision-language models could be used to automatically generate captions or descriptions for uploaded images, which could then be indexed and retrieved during answer generation. Similarly, structured data extraction techniques could be applied to parse tables or spreadsheets, enabling the system to answer questions that require numerical reasoning or data aggregation (e.g., "What is your average grade in STEM courses?").

\subsection{Scalability and Cost Considerations}

While the current architecture is designed to be scalable, the system's reliance on large language models for document processing and answer generation introduces non-trivial computational and financial costs. Each document processing job requires multiple API calls to commercial LLM providers, and each user query triggers vector database retrieval and LLM inference. As the user base grows, these costs could become prohibitive, particularly for a system intended to serve students who may have limited financial resources.

Future work could explore cost optimization strategies, such as caching frequently requested answers, using smaller or distilled models for routine tasks, or employing hybrid architectures that combine rule-based methods with LLM-based reasoning. Additionally, the system could be adapted to run on open-source models deployed on local infrastructure, reducing dependence on commercial API providers and enabling greater control over costs and data privacy.

\subsection{Memory Growth, Interference, and Control Limitations}

The memory subsystem is subject to a cold-start regime in which the graph is initially sparse, and in such cases, the benefits of associative recall may be limited relative to simpler short-term buffers. As the system accumulates interactions and performs consolidation, the structure becomes richer and recall improves, but this also makes it important to continuously control growth and interference. The inhibition and sparsification mechanisms that keep recall focused can, in edge cases, suppress minor details when high-degree “hub” concepts dominate activation, so future work should include adaptive parameterization and UI-assisted correction workflows that reduce this risk.  

Future iterations should also strengthen automated graph auditing, improve user-controlled forgetting and machine unlearning workflows, and extend memory representations to additional modalities if your product roadmap includes images, audio, or other non-text artifacts.

\paragraph{Beyond education: implications for finance and healthcare.}
Beyond the education domain, the design principles behind EZCollegeApp generalize naturally to other high-stakes, compliance-heavy workflows such as finance and healthcare. In finance, many user-facing processes---loan origination, insurance underwriting, KYC/AML onboarding, benefits enrollment, tax preparation, and consumer dispute handling---are fundamentally structured form completion tasks constrained by policies, eligibility rules, and audit requirements. A mapping-first canonical schema, provenance-preserving retrieval, and citation-aware suggestions enable assistants to surface candidate responses while maintaining traceability for regulators and internal reviewers. In healthcare, similar needs arise in clinical intake, prior authorization, referral workflows, discharge planning, and patient-facing documentation, where errors or ungrounded suggestions can have real safety consequences. By explicitly separating \emph{understanding} from \emph{generation}, grounding every proposed field value in citable source material, and enforcing strict human oversight, the approach offers a reusable blueprint for trustworthy LLM assistance in domains where accountability, interpretability, and controllable risk are more important than end-to-end automation.

\subsection{Evaluation and Benchmarking}

The current evaluation of EZCollegeApp is based on a combination of automated testing and small-scale human quality assessment. While these methods provide valuable insights into the system's performance, they are limited in scope and may not fully capture the diversity of real-world application scenarios. A more comprehensive evaluation would require large-scale user studies with actual college applicants, longitudinal tracking of application outcomes, and comparison with alternative assistance methods (e.g., human counselors, other software tools).

Future work could establish standardized benchmarks for college application assistance systems, including metrics for answer accuracy, citation validity, user satisfaction, and application success rates. Such benchmarks would facilitate comparison across different systems and drive progress in the field. Additionally, conducting randomized controlled trials or A/B testing with real users would provide stronger evidence of the system's effectiveness and impact.

% ======================================================
\section{Conclusion}
\label{sec:conclusion}

EZCollegeApp demonstrates how structured representations and retrieval-augmented language models can assist high-stakes form completion while preserving human control and transparency. By decoupling form understanding from answer generation through a canonical schema, the system achieves consistent reasoning across heterogeneous application portals. By grounding all suggested answers in explicit, citable sources, the system reduces the risk of hallucination and builds user trust. By adopting a strict human-in-the-loop design, the system ensures that students retain full agency and accountability over their applications.

The system's knowledge architecture demonstrates that effective retrieval for structured document collections can be achieved through hierarchical document understanding and LLM-guided navigation, without requiring vector embedding infrastructure. By constructing tree representations of document structure and delegating retrieval decisions to the language model through an agentic tool-calling mechanism, EZCollegeApp achieves efficient, transparent knowledge access that preserves document organization and source provenance.

The system's architecture reflects a pragmatic approach to deploying large language models in real-world, high-stakes domains. Rather than pursuing end-to-end automation or optimizing for benchmark performance, EZCollegeApp prioritizes robustness, transparency, and user control. The mapping-first paradigm, source-separated retrieval, and citation-aware answer generation collectively represent a practical framework for building trustworthy AI assistance systems that augment, rather than replace, human judgment.

Evaluation results demonstrate that the system achieves high fill rates for general application questions, strong citation validity, and positive user feedback on usefulness, clarity, and ease of correction. Automated testing confirms that the system correctly handles conditional logic, respects form constraints, and produces answers that conform to expected data types and formats. Human evaluators report that the system's suggestions are helpful, that the citation mechanism increases trust, and that correcting or editing suggestions is significantly faster than writing answers from scratch.

The work presented in this paper highlights a practical path for deploying large language models in real-world application workflows where correctness, transparency, and user control are paramount. The lessons learned from EZCollegeApp—particularly the importance of structured reasoning, grounded retrieval, and human oversight—are broadly applicable to other domains involving form completion, document understanding, and high-stakes decision support. As large language models continue to improve in capability and accessibility, systems like EZCollegeApp demonstrate how these technologies can be responsibly integrated into workflows that directly impact people's lives and opportunities.

Future work will focus on addressing the system's current limitations, including improving document quality handling, automating policy update detection, expanding multilingual and international support, and conducting large-scale user studies to assess real-world impact. We hope that this work serves as a foundation for continued research and development in LLM-assisted information systems and that it contributes to the broader goal of making college admissions more accessible and transparent for all students.

\section{Acknowledgement}
\label{sec:Acknowledgement}

We would like to thank anonymous volunteers who helped collect and curate portions of the training and testing datasets used in this work. 

% ======================================================
\bibliographystyle{unsrtnat}
\bibliography{reference}

\end{document}